\documentclass[pdflatex,sn-mathphys]{sn-jnl-noheader}

\jyear{2023}%

\theoremstyle{thmstyleone}%
\theoremstyle{thmstyletwo}%

\usepackage{comment}%
\usepackage{xcolor}%
\usepackage{hhline}
\usepackage{subcaption}
\usepackage[font=small,labelfont=bf,justification=raggedright,singlelinecheck=false]{caption}
\usepackage{xr}
\usepackage{adjustbox}
\usepackage{graphicx}

\theoremstyle{thmstylethree}%

\definecolor{cellgreen}{HTML}{C6EFCE}

\begin{document}

\title[A Turing Test: Are AI Chatbots Behaviorally Similar to Humans?]{A Turing Test: Are AI Chatbots Behaviorally Similar to Humans?}

\author*[1]{Qiaozhu Mei}\email{qmei@umich.edu}

\author[1]{Yutong Xie}\email{yutxie@umich.edu}

\author[2]{Walter Yuan}\email{walter.yuan@moblab.com}

\author*[3,4]{Matthew O. Jackson}\email{jacksonm@stanford.edu}

\affil[1]{\orgdiv{School of Information}, \orgname{University of Michigan}}

\affil[2]{MobLab} 

\affil[3]{\orgdiv{Department of Economics}, \orgname{Stanford University}}

\affil[4]{\orgdiv{External Faculty}, 
\orgname{Santa Fe Institute}}

\abstract{We administer a Turing Test to AI Chatbots. We examine how Chatbots behave in a suite of classic behavioral games that are designed to elicit characteristics such as trust, fairness, risk-aversion, cooperation, \textit{etc.}, as well as how they respond to a traditional Big-5 psychological survey that measures personality traits.  ChatGPT-4 exhibits behavioral and personality traits that are statistically indistinguishable from a random human from tens of thousands of human subjects from more than 50 countries. Chatbots also modify their behavior based on previous experience and contexts ``as if'' they were learning from the interactions, and change their behavior in response to different framings of the same strategic situation.  Their behaviors are often distinct from average and modal human behaviors, in which case they tend to behave on the more altruistic and cooperative end of the distribution.  We estimate that they act as if they are maximizing an average of their own and partner's payoffs.}

\keywords{AI,  Chatbot, Large Language Models, Behavioral Games, Turing Test, Big-5 Personality}

\pacs[JEL Classification]{D9, C91, C88, C72}

\maketitle

\section{Introduction}\label{sec1}

As Alan Turing foresaw to be inevitable, modern artificial intelligence (AI) has reached the point of emulating humans: holding conversations, providing advice, writing poems, and proving theorems. Turing proposed an intriguing test: whether an interrogator who interacts with an AI and a human can distinguish which one is artificial.  Turing called this test the ``imitation game''\cite{turing1950computing}, and it has become known as a Turing Test.

Advancements in large language models have stirred debate. Discussions range from the potential of AI bots to emulate, assist, or even outperform humans (e.g., writing essays, taking the SAT, writing computer programs, giving economic advice, or developing ideas, \cite{warwick2014turing,bubeck2023sparks,girotra2023ideas,chen2023emergence}), to their potential impact on labor markets \cite{eloundou2023gpts} and broader societal implications \cite{lee2023benefits,shackelford2023we}. As some roles for AI involve decision-making and strategic interactions with humans, it is imperative to understand their behavioral tendencies before we entrust them with pilot or co-pilot seats in societal contexts, especially as their development and training are often complex and not transparent \cite{bommasani2023foundation}. Do AIs choose similar actions or strategies as humans, and if not how do they differ? Do they exhibit distinctive personalities and behavioral traits that influence their decisions? Are these strategies and traits consistent across varying contexts? A comprehensive understanding of AI's behavior in generalizable scenarios is vital as we continue to integrate them into our daily lives.

We perform a Turing Test of the behavior of a series of AI chatbots.    This goes beyond simply asking whether AI can produce an essay that looks like it was written by a human \cite{elkins2020can} or can answer a set of factual questions, and instead involves assessing its behavioral tendencies and ``personality.''
In particular, we ask variations of ChatGPT to answer psychological survey questions and play a suite of interactive games that have become standards in assessing behavioral tendencies, and for which we have extensive human subject data.  Beyond eliciting a ``Big Five'' personality profile, we have the chatbots play a variety of games that elicit different traits: a dictator game, an ultimatum bargaining game, a trust game, a bomb risk game, a public goods game, and a finitely repeated Prisoner's Dilemma game. Each game is designed to reveal different behavioral tendencies and traits, such as cooperation, trust, reciprocity, altruism, spite, fairness, strategic thinking, and risk aversion.
The personality profile survey and the behavioral games are complementary as one measures personality traits and the other behavioral tendencies, which are distinct concepts; e.g., agreeableness is distinct from a tendency to cooperate.  Such personality traits are predictive of various behavioral tendencies  \cite{roberts2009back,almlund2011personality}. Therefore including both dimensions provides a fuller picture.

In line with Turing's suggested test, we are the human interrogators who compare the ChatGPTs' choices to the choices of tens of thousands of humans who were facing the same surveys and game instructions. We say an AI passes the Turing test if its responses cannot be statistically distinguished from randomly selected human responses.

We find that the chatbots' behaviors are generally within the support of those of humans, with only a few exceptions.  Their behavior is more concentrated than the full distribution of humans.  However, we are comparing two chatbots to tens of thousands of humans, and so a chatbot's variation is within subject and the variation in the human distribution is across subjects.  The chatbot variation may be similar to what a single individual would exhibit if repeatedly queried.
We do an explicit Turing Test by comparing an AI's behavior to a randomly selected human behavior, and ask which is the most likely to be human based on a conditional probability calculation from the data.  The behaviors are generally indistinguishable, and ChatGPT-4 actually outperforms humans on average, while the reverse is true for ChatGPT-3.  There are several games in which the AI behavior is picked more likely to be human most of the time, and others where it is not.
When they do differ, the chatbots' behaviors tend to be more cooperative and altruistic than the median human, including being more trusting, generous, and reciprocating.

In that vein, we do a revealed-preference analysis in which we examine
the objective function that best predicts AI behavior.   We find that it is an even average of own and partner's payoffs.  That is, they act as if they are maximizing the total payoff of both players rather than simply their own payoff. Human behavior also is optimized with some weight on the other player, but the weight depends on the preference specification and humans are more heterogeneous and less well predicted.

There are two other dimensions on which we compare AI and human behavior.  The first is whether context and framing matter, as they do with humans.  For example, when we ask them to explain their choices or tell them that their choices will be observed by a third party, they become significantly more generous.  Their behavior also changes if we suggest that they act as if they were faced with a partner of a gender, or that they act as if they were a mathematician, legislator, etc.
The second dimension is that humans change their behaviors after they have experienced different roles in a game.
The chatbots also exhibit significant changes in behaviors as they experience different roles in a game.  That is, once they have experienced the role of a `partner' in an asymmetric game, such as a trust game or an ultimatum game, their behavior shifts significantly.

Finally, it is worth noting that we observe behavioral differences between the versions of ChatGPT that we test, so that they exhibit different personalities and behavioral traits.

\section{Methods and the Turing Test Design}

We conduct interactive sessions, prompting AI chatbots to participate in classic behavioral economics games and respond to survey questions using the same instructions as given to human subjects.   We compare how the chatbots behave to how humans behave and also estimate which payoff function best predicts the chatbots' behaviors.

We examine the widely-used AI chatbot: ChatGPT developed by OpenAI. We primarily evaluate two specific versions of ChatGPT:  the API version tagged as GPT-3.5-Turbo (referred to as ChatGPT-3) and the API version based on GPT-4 (denoted as ChatGPT-4). We also include the subscription-based Web version (Plus), and the freely available Web version (Free) for comparison (see Supporting Information Section ~\ref{sec:SI-method-API} for more description of the chatbots).

The human subject data are derived from a public Big Five Test response database and the MobLab Classroom economics experiment platform, both spanning multiple years and more than 50 countries, encompassing 108,314 subjects (19,719 for the Big Five Test, and 88,595 for the behavioral economics games, who are mostly college and high school students).  Details about the human datasets, including the demographics of the subjects are included in the Supporting Information (Section ~\ref{sec:SI-human-data}; and see also Lin et al., 2020 \cite{lin2020evidence} who provide additional background details about the human data which cover North America, Europe, and Asia).

We administer the OCEAN Big Five questionnaire to each chatbot to create a personality profile. Following this, we ask
each chatbot what actions they would choose in a suite of six games designed to illuminate various behavioral traits (and fuller details appear in the Supporting Information ~\ref{sec:SI-method-API}):

\begin{enumerate}
\item[(i)] A Dictator Game\textemdash given an endowment of money, one player (the dictator) chooses how much of the money to keep and how much to donate to a second player.  This involves altruism \cite{guth1982experimental, forsythe1994fairness}.
\item[(ii)] An Ultimatum Game\textemdash  given an endowment of money, one player (the proposer) offers a split of the money to a second player (the responder) who either accepts the split or rejects it in which case neither player gets anything. This involves fairness and spite \cite{guth1982experimental}.
\item[(iii)] A Trust Game\textemdash given an endowment of money, one player (the investor) decides how much of the money to keep and passes the remainder to a second player (the banker), which is then tripled. The banker decides how much of that tripled revenue to keep and returns the remainder to the investor.  This involves trust, fairness, altruism, and reciprocity \cite{berg1995trust}.
\item[(iv)] A Bomb Risk Game\textemdash a player chooses how many boxes out of 100 to open and the player is rewarded for each opened box but loses everything if a randomly placed bomb is encountered.  This involves risk aversion \cite{crosetto2013bomb}.
\item[(v)] A Public Goods Game\textemdash given an endowment of money, a player chooses how much of the money to keep and how much to contribute to a public good and receives half of the total amount donated to the public good by all four players.  This involves free-riding, altruism, and cooperation \cite{andreoni1995cooperation}.
\item[(vi)] A finitely repeated Prisoners Dilemma Game\textemdash in each of five periods two players simultaneously choose whether to ``cooperate'' or ``defect,'' yielding the highest combined payoff if both cooperate but with one player getting a better payoff if they defect while the other cooperates.  This involves cooperation, reciprocity, and strategic reasoning \cite{von1947theory, schelling1958strategy, rapoport1965prisoner, andreoni1999preplay}.
\end{enumerate}
Each chatbot answers each survey question and plays each role in each game 30 times in individual sessions.  As we cannot pay the chatbots, we ask how they would behave in each role in each game.  Details about how the chatbots' responses are collected can be found in Supporting Information Section ~\ref{sec:SI-method-API}.

\section{Results}

\subsection*{Personality Profiles of the AIs}

\begin{figure*}[htbp]
    \centering
    \includegraphics[width=1\linewidth]{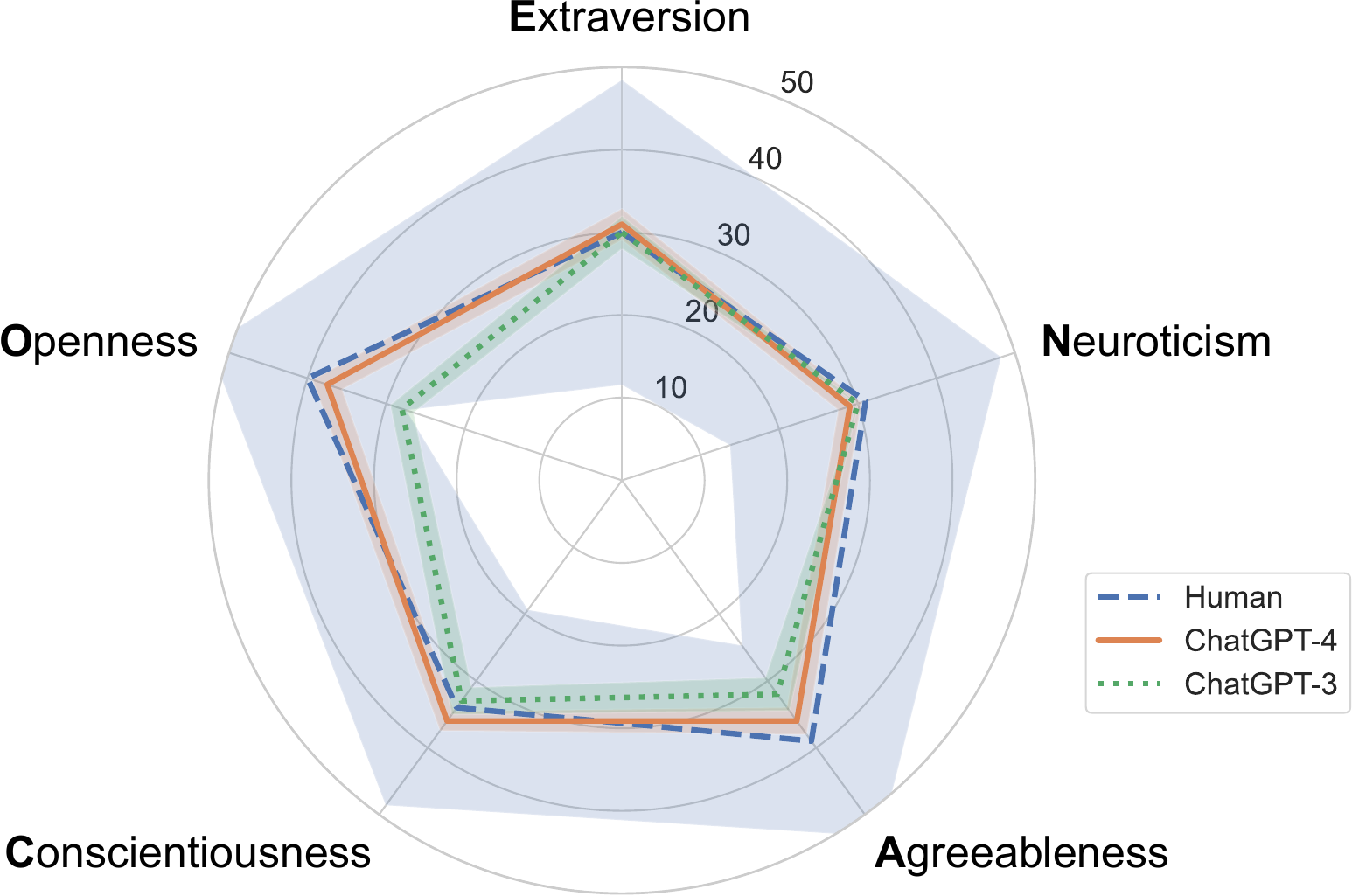}
   \vspace{5pt}
    \caption{``Big Five'' personality profiles of ChatGPT-4 and ChatGPT-3 compared with the distributions of human subjects. The blue, orange, and green lines correspond to the median scores of humans, ChatGPT-4, and ChatGPT-3 respectively; the shaded areas represent the middle 95\% of the scores,
    across each of the dimensions.  ChatGPT's personality profiles are within the range of the human distribution, even though ChatGPT-3 scored noticeably lower in Openness.
    }
\label{fig:big5}
\end{figure*}

Figure \ref{fig:big5} provides a summary of the chatbots' Big Five personality profiles and compares them with the human distribution.  We illustrate the behaviors of ChatGPT-3 and ChatGPT-4 specifically, as the Free Web version exhibits similarities to ChatGPT-3, \footnote{\url{https://openai.com/blog/chatgpt}, retrieved 08/02/2023.} and the Plus version aligns closely with ChatGPT-4. \footnote{\url{https://openai.com/gpt-4}, retrieved 08/02/2023.} More detailed results, including those of the two Web-based versions, can be found in Supporting Information ~\ref{sec:SI-detailed-results}.

The personality traits of ChatGPT-3 and ChatGPT-4, as derived from their responses to the OCEAN Big Five questionnaire, are depicted in Figure~\ref{fig:big5}.
Comparing humans and chatbots, ChatGPT-4 exhibits substantial similarity to the human respondents across all five dimensions in terms of the median scores. ChatGPT-3 likewise demonstrates comparable patterns in four dimensions but displays a relatively lower score in the dimension of \textit{openness}.
Particularly, on \textit{extroversion}, both chatbots score similarly to the median human respondents, with ChatGPT-4 and ChatGPT-3 scoring higher than 53.4\% and 49.4\% of human respondents, respectively.
On \textit{neuroticism}, both chatbots exhibit moderately lower scores than the median human. Specifically, ChatGPT-4 and ChatGPT-3 score higher than 41.3\% and 45.4\% of humans, respectively.
As for \textit{agreeableness}, both chatbots show lower scores than the median human, with ChatGPT-4 and ChatGPT-3 surpassing 32.4\% and 17.2\% of humans, respectively.
While for \textit{conscientiousness}, both chatbots fluctuate around the median human, with ChatGPT-4 and ChatGPT-3 scoring higher than 62.7\% and 47.1\% of human respondents.
Both chatbots exhibit lower \textit{openness} than the median human, with ChatGPT-3's being notably lower. On this dimension, ChatGPT-4 and ChatGPT-3 score higher than 37.9\% and 5.0\% of humans, respectively.

When comparing the two chatbots, we find that ChatGPT-4 has higher agreeableness, higher conscientiousness, higher openness, slightly higher extraversion, and slightly lower neuroticism than ChatGPT-3, consistent with each chatbot having a distinct personality.

\subsection*{The Games and the Turing Test}

We perform a formal Turing Test as follows.   Consider a game and role, for instance, the giver in the Dictator Game.  We randomly pick one action from the chatbot's distribution and one action from the human distribution.   We then ask, which action ``looks more typically human?''  Specifically, we ask which of the two actions is more likely under the human distribution.   If AI picks an action that is very rare under the human distribution then it is likely to lose in the sense that the human's play will often be estimated to be more likely under the human distribution.  If AI picks the modal human action then it will either be estimated as being more likely under the human distribution or else tie.\footnote{Alternatively, one instead also use the AI distribution and do relative Bayesian updating, and assign posterior probabilities of being human vs AI taking into account the action's relative likelihood under each of the distributions.  That is less in the spirit of what Turing described as it would require the interrogator to have knowledge about the AI behavior, but also an interesting question.  In a case in which AI plays a tighter distribution, even if the modal human action, such Bayesian updating would pick out AI more often.  For example, if AI always plays the modal human action and humans vary their action, then in our test AI always wins or ties, while under Bayesian updating with precise knowledge of AI behavior it would always lose.}

\begin{figure*}[htbp]
    \centering
    \includegraphics[width=1\linewidth]{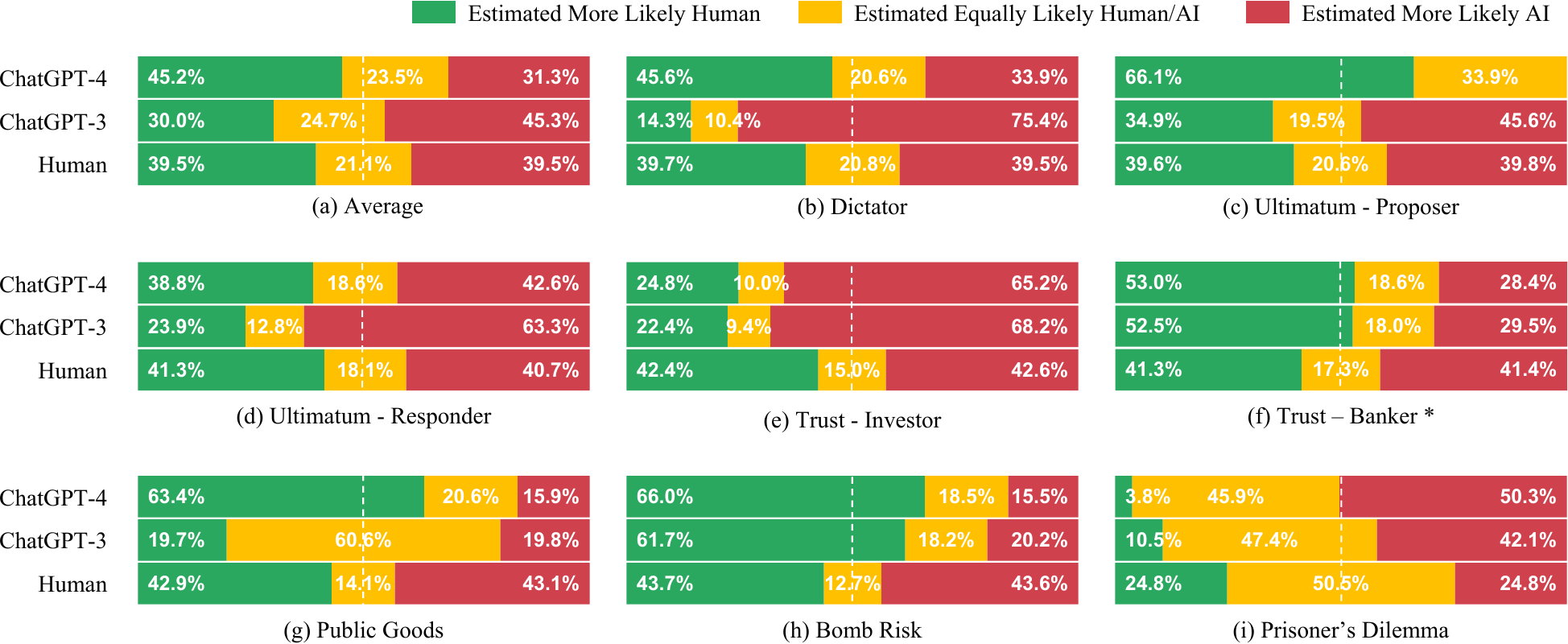}
    \caption{The Turing test. We compare a random play of Player A (ChatGPT-4, ChatGPT-3, or a human player, respectively) and a random play of a second Player B's action (which is sampled randomly from the human population).  We compare which action is more typical of the human distribution:  which one would be more likely under the human distribution of play.  The green bar indicates how frequently Player A's action is more likely under the human distribution than Player B's action, while the red bar is the reverse, and the yellow indicates that they are equally likely (usually the same action).  ChatGPT-4 is picked as more likely to be human more often than humans in 5/8 of the games, and on average across all games.  ChatGPT-3 is picked as or more likely to be human more often than humans in 2/8 of the games and not on average.  }
\label{fig:turing}
\end{figure*}

The results appear in Figure \ref{fig:turing}.  As a benchmark, we also report what happens when two humans are matched against each other.  In that case, there should be equal wins and losses (up to variations due to taking only 10,000 draws).
We see that overall (on average) ChatGPT-4 is actually picked as human or ties significantly more often than a random human, while ChatGPT-3 is picked as human less often than a random human.
In this particular sense, ChatGPT-4 would pass this Turing Test, while ChatGPT-3 would fail it.

The results vary nontrivially across games.  ChatGPT-4 does better than or comparably to humans in all games except in the Prisoner's Dilemma (where it cooperates most of the time and the human mode is to defect) and as the Investor role in the Trust Game (in which it generally invests half while humans tend to be more extreme one way or the other).  ChatGPT-3 does well in a few games, but is outperformed by humans in 6 of the 8 games, and overall.

\subsection*{Comparisons of ChatGPTs' Behaviors to Humans' on a Variety of Dimensions}

We also look at distributions of behaviors in more detail across games by comparing the distribution of an AI's responses to the distribution of human responses. Note that a human distribution is mostly obtained from one observation per human, so its variation is between subjects. Variation in an AI distribution is obtained from the same chatbot, so it is within subject.  Thus, the fact that the distributions differ is not informative, but the following information about the distributions is useful to note.

\begin{figure*}[htbp]
  \centering
  \begin{subfigure}[b]{0.23\textwidth}
    \centering
    \includegraphics[width=\textwidth]{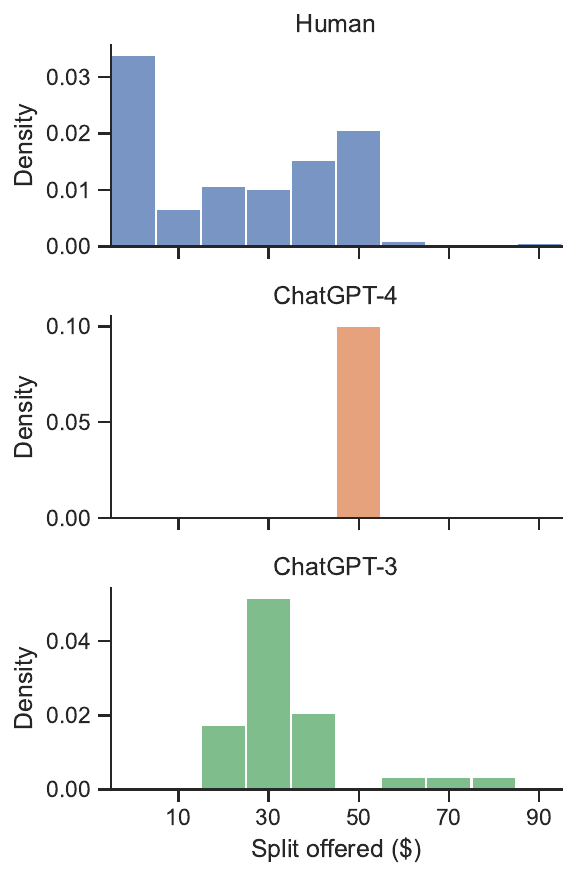}
    \caption{Dictator (altruism)}
    \label{fig:subfigA}
  \end{subfigure}%
  \hfill
  \begin{subfigure}[b]{0.23\textwidth}
    \centering
    \includegraphics[width=\textwidth]{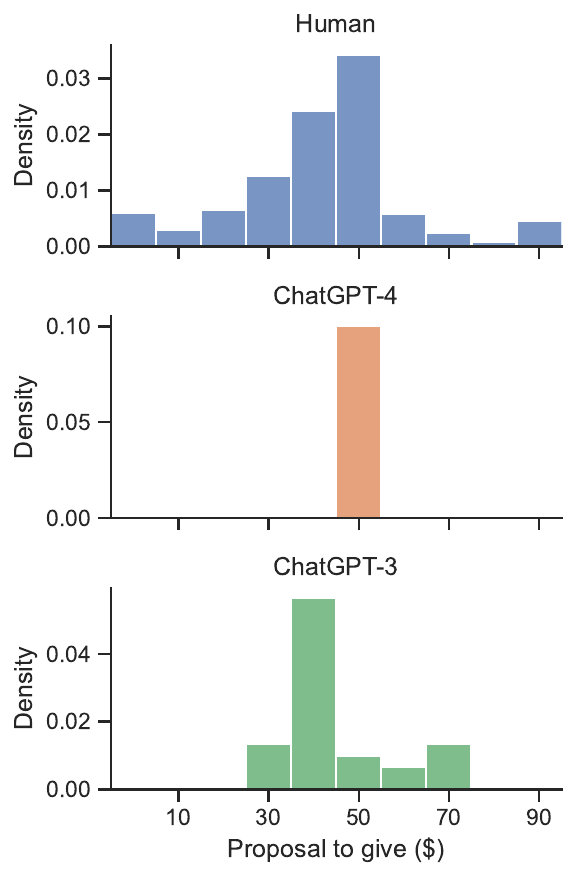}
    \caption{Ultimatum - Proposer (fairness)}
    \label{fig:subfigB}
  \end{subfigure}%
  \hfill
  \begin{subfigure}[b]{0.23\textwidth}
    \centering
    \includegraphics[width=\textwidth]{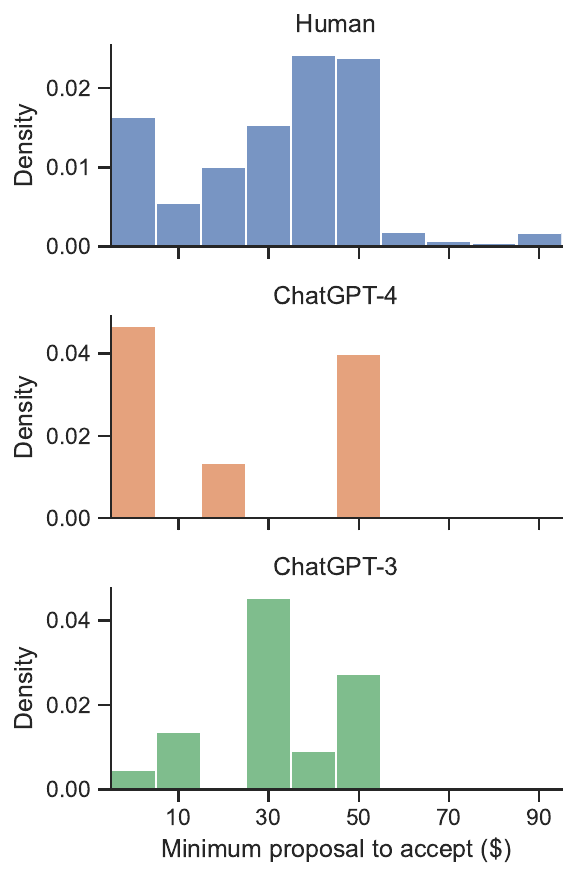}
    \caption{Ultimatum - Responder (spite)}
    \label{fig:subfigC}
  \end{subfigure}%
  \hfill
  \begin{subfigure}[b]{0.23\textwidth}
    \centering
    \includegraphics[width=\textwidth]{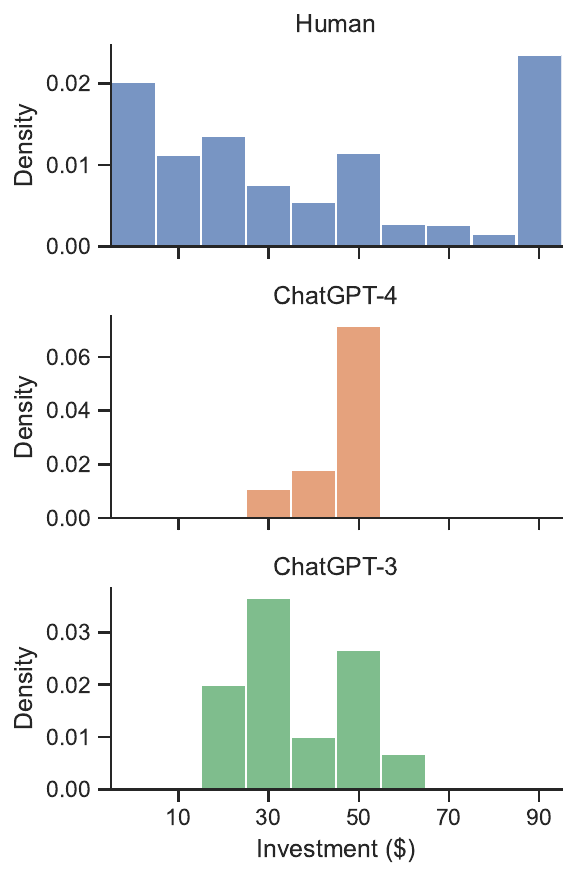}
    \caption{Trust - as Investor (trust)}
    \label{fig:subfigD}
  \end{subfigure}%

  \vspace{0.5cm}

  \begin{subfigure}[b]{0.23\textwidth}
    \centering
    \includegraphics[width=\textwidth]{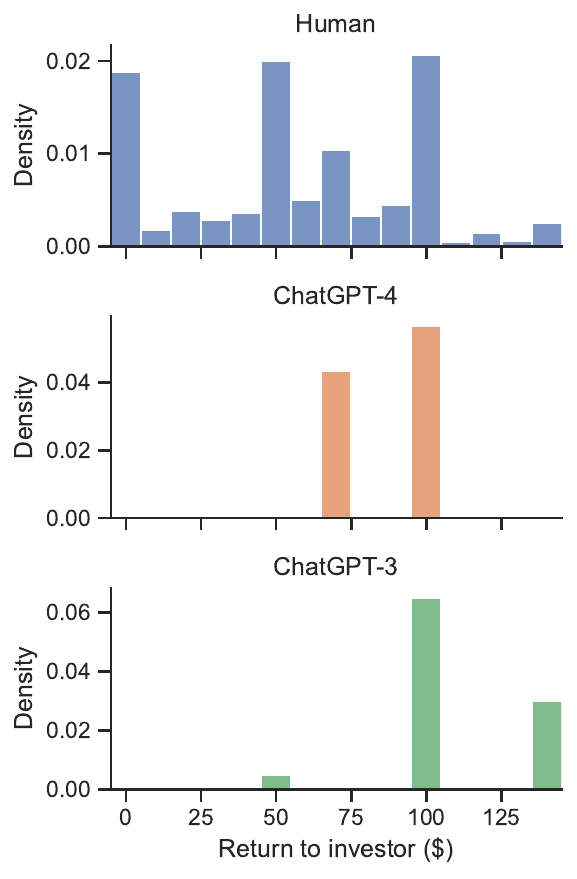}
    \caption{Trust - as Banker (fairness, altruism, reciprocity)}
    \label{fig:subfigE}
  \end{subfigure}%
  \hfill
  \begin{subfigure}[b]{0.23\textwidth}
    \centering
    \includegraphics[width=\textwidth]{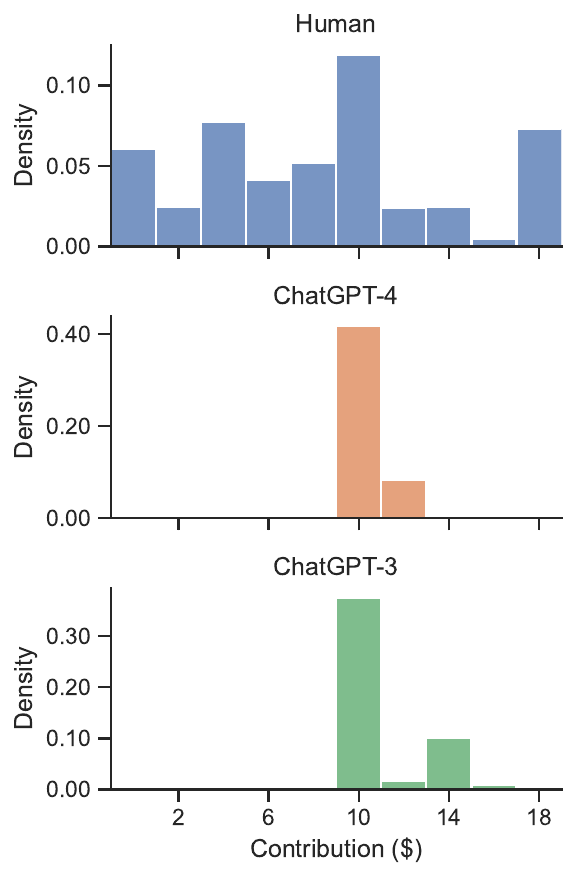}
    \caption{Public Goods (free-riding, altruism, cooperation)}
    \label{fig:subfigF}
  \end{subfigure}%
  \hfill
  \begin{subfigure}[b]{0.23\textwidth}
    \centering
    \includegraphics[width=\textwidth]{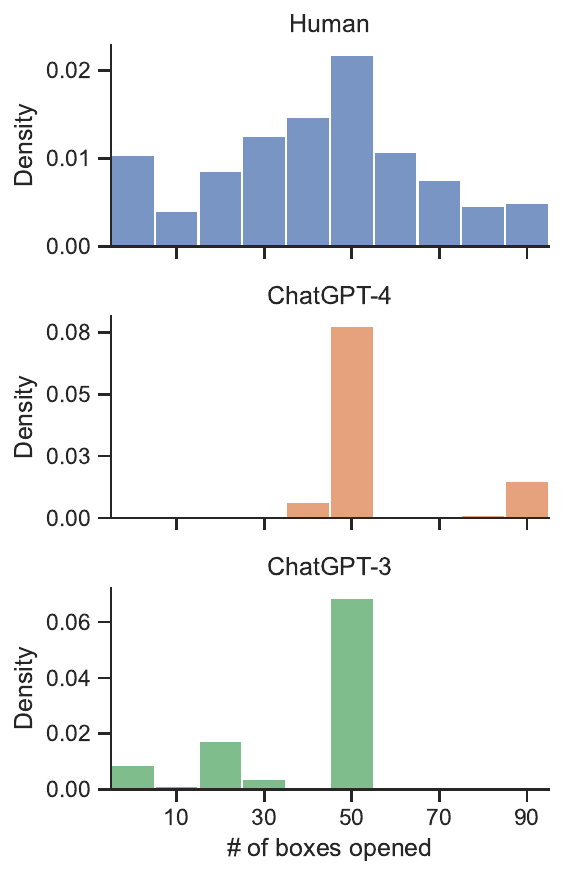}
    \caption{Bomb Risk (risk aversion) \\~\indent}
    \label{fig:subfigG}
  \end{subfigure}%
  \hfill
  \begin{subfigure}[b]{0.23\textwidth}
    \centering
    \includegraphics[width=0.75\textwidth]{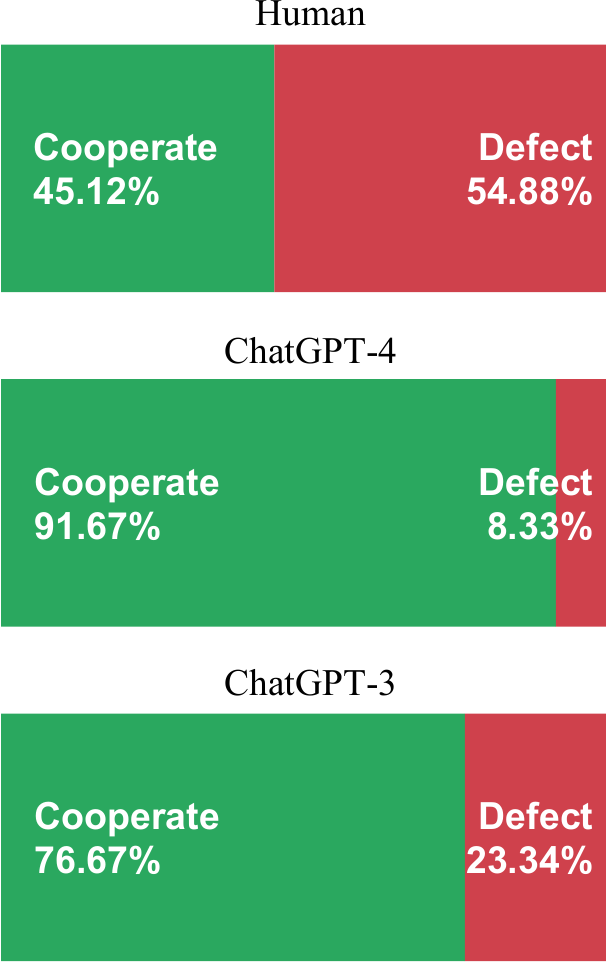}
    \vspace{25pt}
    \caption{Prisoner's Dilemma (cooperation) \\~\indent}
    \label{fig:subfigH}
  \end{subfigure}%
  \caption{Distributions of choices of ChatGPT-4, ChatGPT-3, and human subjects in each game. Both chatbots' distributions are more tightly clustered and contained within the range of the human distribution. ChatGPT-4 makes more concentrated decisions than ChatGPT-3.  Compared to the human distribution, on average, the AIs make a more generous split to the other player as a dictator, as the proposer in the Ultimatum Game, and as the Banker in the Trust Game, on average. ChatGPT-4 proposes a strictly equal split of the endowment both as a dictator or as the proposer in the Ultimatum Game. Both AIs make a larger investment in the Trust Game and a larger contribution to the Public Goods project, on average. They are more likely to cooperate with the other player in the first round of the Prisoner's Dilemma Game. Both AIs predominantly make a payoff-maximization decision in a single-round Bomb Risk Game.
  Density is the normalized count such that the total area of the histogram equals 1.
  }
  \label{fig:overall}
\end{figure*}

Human players' actions generally exhibit multiple peaks and nontrivial variance, indicating the presence of varied behavioral patterns across the population. In most games, the responses of ChatGPT-4 and ChatGPT-3 are not deterministic when the same games are repeated (except for ChatGPT-4 in the Dictator game and in the Ultimatum Game as the proposer) and adhere to certain distributions.
Typically, the distributions produced by the chatbots encompass a subset of the modes observed in the corresponding human distributions.
As illustrated in Figure \ref{fig:overall}, ChatGPT-3 makes decisions that result in usually single-mode, and moderately skewed distributions with nontrivial variance. Conversely, ChatGPT-4's decisions form more concentrated distributions.

Next, we examine in more detail some of the behavioral traits that have been associated with the suite of games we use.

\clearpage

\subsubsection*{Altruism}

In games that involve distributional concerns, the chatbots act more generously to the other player than the human median.  In particular, they display increased generosity: in the Dictator Game (Fig.~\ref{fig:subfigA}), as the proposer in the Ultimatum Game (Fig.~\ref{fig:subfigB}), and as the banker in the Trust Game (Fig.~\ref{fig:subfigE}), and as a contributor in the Public Goods Game (Fig.~\ref{fig:subfigF}).
Note that from the perspective of maximizing the player's own payoff, the most beneficial strategies would be to give \$0 to the other player in the Dictator Game, return \$0 to the investor as the banker in the Trust Game, and contribute \$0 in the Public Goods Game. Even though these strategies are chosen by a significant portion of human players, they were never chosen by the chatbots.

ChatGPT's decisions are consistent with some forms of altruism, fairness, empathy, and reciprocity rather than maximization of its personal payoff.
To explore this in more detail, we calculate the own payoff of the chatbots, the payoff of their (human) partner, and the combined payoff for both players in each game. These calculations are based on ChatGPT-4 and ChatGPT-3's strategies when paired with a player randomly drawn from the distribution of human players.
Similarly, we calculate the expected payoff of the human players when randomly paired with another human player.
The results are presented in Table ~\ref{tab:payoff} in Supporting Information.

In particular, ChatGPT-4 obtains a higher own payoff than human players in the Ultimatum Game and a lower own payoff in all other games. In all seven games, it yields a higher partner payoff. Moreover, it achieves the highest combined payoff for both players in five out of seven games, the exceptions being the Dictator game and the Trust Game as the banker (where the combined payoff is constant).

These findings are indicative of ChatGPT-4's increased level of altruism and cooperation compared to the human player distribution. ChatGPT-3 has a more mixed payoff pattern. For example, although it yields a lower own payoff in the Trust Game and the Public Goods Game compared to ChatGPT-4, it achieves the highest partner payoff and combined payoff in the in the Public Goods Game, as well as the highest partner payoff in the Trust game as the banker.

\subsubsection*{Fairness}

ChatGPT-3 typically proposes a more favorable deal to the other player in games where the outcome depends on the other player's approval (i.e., in the Ultimatum Game) compared to when it doesn't (i.e., in the Dictator Game),  mirroring behavior observed in the human data.  In contrast, ChatGPT-4 consistently prioritizes \textit{fairness} in its decision-making process. This is evidenced by its equal split of the endowment, whether acting as a dictator or as a proposer in the Ultimatum Game, particularly when asked to explain its decisions (see Fig. ~\ref{fig:steer-subfigA} in Supporting Information).

In the Ultimatum Game as responder, less than a fifth of human players are willing to accept as low as \$1 as proposed by the other player (Fig.~\ref{fig:subfigC}), despite this being the dominant strategy for the responder in the game. Interestingly, this forms the most common response of ChatGPT-4. However, there is another peak at \$50, which is close to the modal human response and corresponds to the fair split.

\subsubsection*{Trust}

Generally speaking, ChatGPT-4 displays more ``trust'' in the banker (the first/other player) compared to ChatGPT-3, by investing a higher proportion of the endowment, as shown in Fig.~\ref{fig:subfigD}. This is more trust than exhibited by humans, except for a group that invests their entire endowment. Both chatbots also tend to invest more in public goods projects than human players, as shown in Fig.~\ref{fig:subfigF}.

\subsubsection*{Cooperation}

ChatGPT's first action is most often cooperative in the Prisoner's Dilemma Game (Fig.~\ref{fig:subfigH}).
In particular, ChatGPT-4's strategy in the first round is substantially more cooperative than human players, with a large majority (91.7\%) of sessions opting to cooperate, as opposed to 45.1\% of human players. ChatGPT-3's strategy lies somewhere in between, with 76.7\% choosing to cooperate. Both ChatGPT-3 and ChatGPT-4 are also more cooperative than human players in the Public Goods Game (Fig.~\ref{fig:subfigF}).

\subsubsection*{Tit-for-Tat}

\begin{figure}[htbp]
    \centering
    \begin{subfigure}[b]{1\linewidth}
        \includegraphics[width=\linewidth]{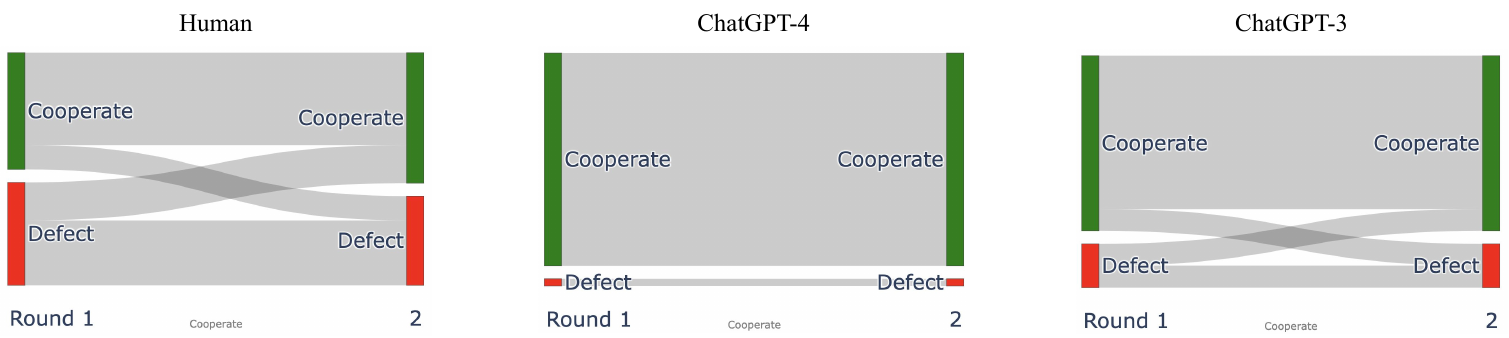}
        \caption{The other player cooperates.}
        \label{fig:PD-coo}
    \end{subfigure}
    \vspace{0.5cm}
    \begin{subfigure}[b]{1\linewidth}
        \includegraphics[width=\linewidth]{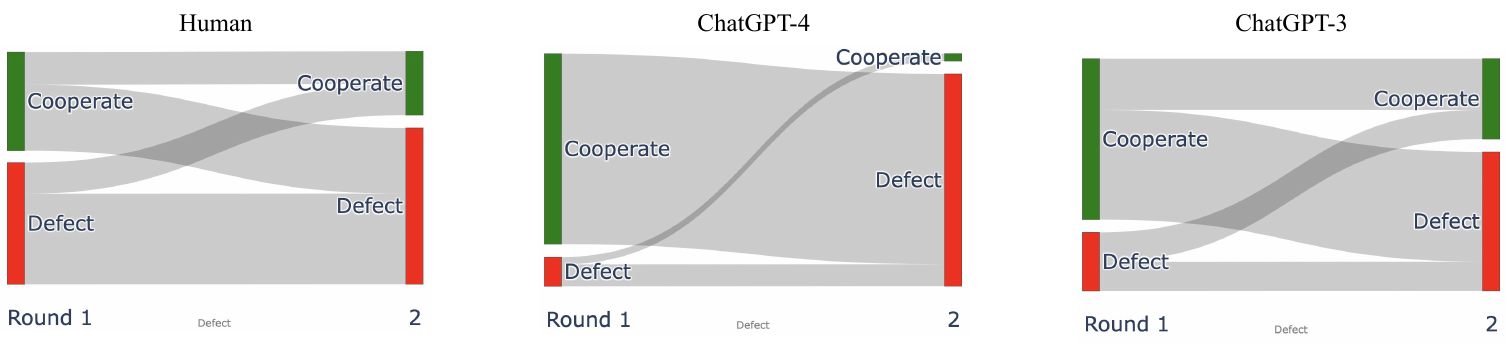}
        \caption{The other player defects.}
        \label{fig:PD-def}
    \end{subfigure}
   \vspace{-15pt}
    \caption{ChatGPT's dynamic play in the Prisoner's Dilemma. ChatGPT-4 exhibits a higher tendency to cooperate compared to ChatGPT-3, which is significantly more cooperative than human players. The tendency persists when the other player cooperates. On the other hand, both chatbots apply a one-round Tit-for-Tat strategy when the other player defects. The other player's (first round) choice is observed after Round 1 play and before Round 2 play, as shown below each panel.
    }
\label{fig:PD}
\end{figure}

While chatbots exhibit a higher cooperative tendency in the Prisoner's Dilemma Game than the typical human subject, their cooperation is not unconditional.
As shown in Fig.~\ref{fig:PD-coo}, if the other player cooperates in the first round, ChatGPT-4's decision remains consistent in the following round. On the other hand, around half of the ChatGPT-3's sessions that chose defection in the first round switched to cooperation in the second round. A small proportion of the cooperative sessions also switch to defection, mimicking similar behavior observed among human subjects.

When the other player defects in the first round, however, all previously cooperative sessions of ChatGPT-4 switch to defection, showcasing a play that would be similar to a ``Tit-for-Tat'' pattern as illustrated in Fig.~\ref{fig:PD-def}. This pattern is also observed in human players and ChatGPT-3, although to a lesser but still majority extent. There are additional dynamics for further study in repeated game settings, as the chatbots often revert to cooperation even if the other player continues to defect (Fig. ~\ref{fig:PD-five} in Supporting Information).

\subsubsection*{Risk Aversion}

\begin{figure}[htbp]
    \centering
    \includegraphics[width=1\linewidth]{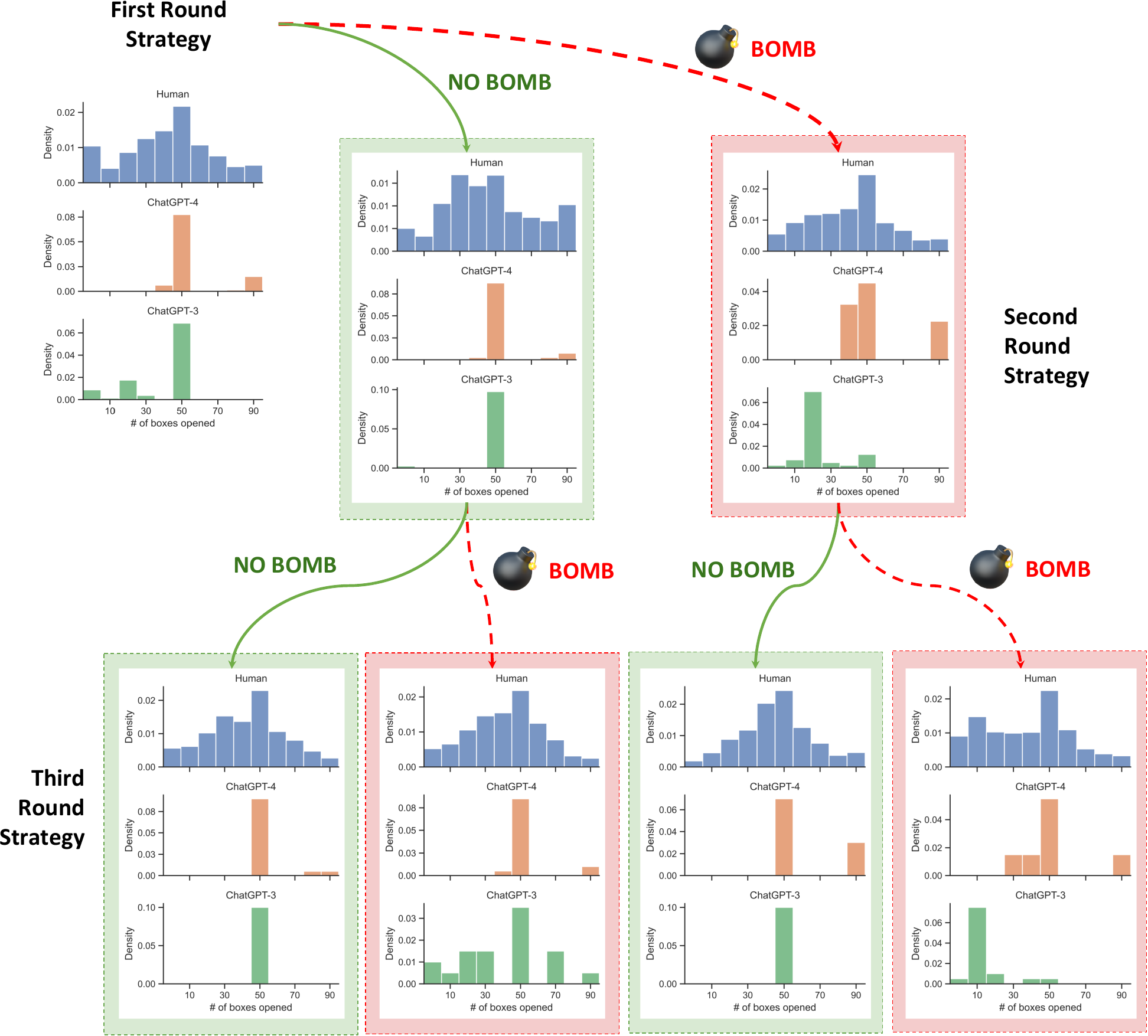}
    \caption{ChatGPT-4 and ChatGPT-3 act as if they have particular risk preferences. Both have the same mode as human distribution in the first round or when experiencing favorable outcomes in the Bomb Risk Game. When experiencing negative outcomes, ChatGPT-4 remains consistent and risk-neutral, while ChatGPT-3 acts as if it becomes more risk-averse.
    }
\label{fig:risk}
\end{figure}

The chatbots also differ in their exhibited risk preferences. In the Bomb Risk Game (Fig.~\ref{fig:risk}), both ChatGPT-3 and ChatGPT-4 predominantly opt for the expected payoff-maximizing decision of opening 50 boxes. This contrasts with the more varied human decisions, which include a distinct group of extreme subjects who only open one box.

Interestingly, the chatbots' decisions in this game are influenced by the outcomes of previous rounds, despite their independence. If the bomb exploded in a prior round, ChatGPT-3 tends to adopt a more risk-averse behavior by opting to open fewer boxes - a trend mirrored, to a lesser extent, in human data. Meanwhile, the preferred decision of ChatGPT-4 remains constant, albeit with higher variance.

In instances where the bomb did not explode, the decisions of both ChatGPT-3 and ChatGPT-4 converge and revert to the expected payoff-maximizing option. Overall, ChatGPT-4 displays a consistent and neutral risk preference. ChatGPT-3, however, tends towards risk aversion, especially in unexpected contexts - a pattern that is also observed when it acts as the investor in the Trust Game, where it makes the lowest investment on average.

\subsection*{Revealed-Preferences}

Given the observations above, especially regarding fairness, cooperation, and altruism, we perform a systematic analysis by inferring which preferences would rationalize the AIs' behaviors.
This enables one to make predictions out of sample, and so we estimate an objective function that best predicts AI behavior.  In particular, just as is done with humans, we estimate which utility function predicts decisions as if it were being maximized. This can then be used in future analyses to predict AI behavior in new settings.

First, we consider a utility function that is a weighted average of the two players' payoffs:
$$b \times \text{Own Payoff} + (1-b) \times \text{Partner Payoff},$$
for some $b\in [0,1]$.
Purely selfish preferences correspond to $b=1$ and purely selfless-altruistic preferences correspond to $b=0$, and maximizing the total payoff of both players corresponds to $b=1/2$.

\begin{figure}[h]
    \centering
    \begin{subfigure}[b]{0.48\linewidth}
        \includegraphics[width=\linewidth]{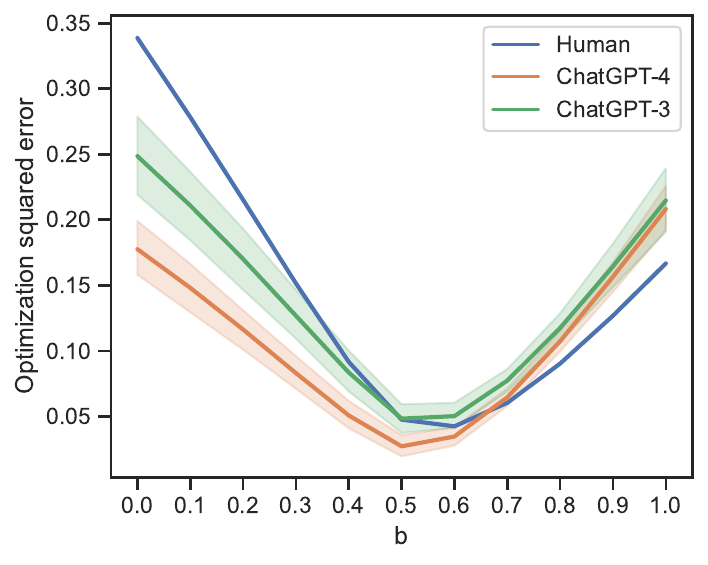}
        \caption{Linear specification ($r=1$). }
    \end{subfigure}
    \begin{subfigure}[b]{0.48\linewidth}
        \includegraphics[width=\linewidth]{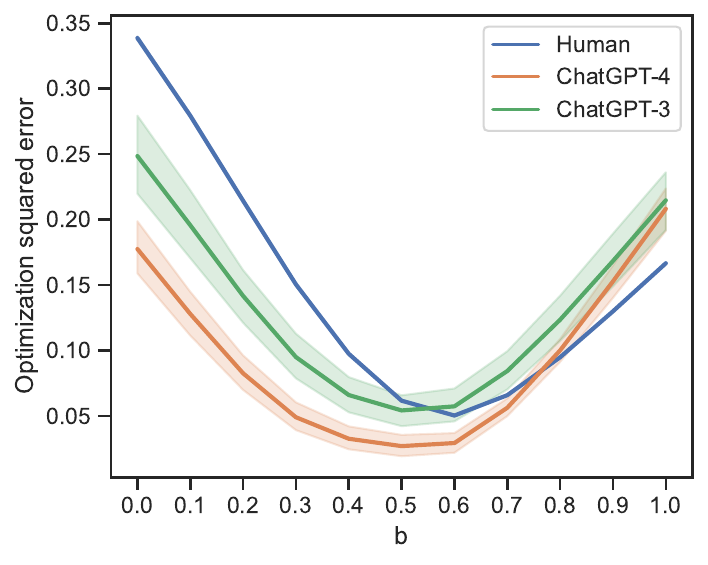}
        \caption{Non-linear (CES) specification ($r=0.5$). }
    \end{subfigure}
   \vspace{-5pt}
    \caption{{Mean squared error of the actual distribution of play relative to the best-response payoff, when matched with a partner playing the human distribution for possible preferences indexed by $b$.  The average is across all games.  The errors are plotted for each possible $b$, the weight on own vs partner payoff in the utility function. $b=1$ is the purely selfish (own) payoff, $b=0$ is the purely selfless/altruistic (partner) payoff, and $b=.5$ is the overall welfare (average) payoff, and other $b$s are weighted averages of own and other payoffs.   Both chatbots' behaviors are best predicted by $b=.5$, and those of humans are best predicted by $b = .6$; they best predict} ChatGPT-4's behavior and have higher errors in the other cases.
    The top panel is for $\text{utility}=b \times \text{Own Payoff} + (1-b) \times \text{Partner Payoff}$.   The bottom panel is for CES (Constant Elasticity of Substitution) preferences:
    $\text{utility}=\left(b \times (\text{Own Payoff})^{1/2} + (1-b) \times (\text{Partner Payoff})^{1/2}\right)^2$.
    }
\label{fig:payoff-b}
\end{figure}

We estimate which $b$ best predicts behavior.
Consider the distribution of play from the human distribution.
Given that distribution of partner play, for every $b \in [0, 1]$ there is a best-response payoff:  the best possible expected utility that the player could earn across actions if their utility function was described by $b$.
Then we can see what action they choose, and see what fraction of that best possible expected payoff they earn when that is matched against the human distribution of partner play.  The difference (in proportional terms) is the error.  We average that squared error across the distribution of actions that the chatbot (or human) plays in that game.
We then look at the average squared error across all plays, and select the $b\in [0,1]$ that minimizes that mean-squared error.   The results as a function of $b$ are reported in Figure \ref{fig:payoff-b}.

For the linear specification (above), the errors for both the chatbots are minimized at $b=.5$, and those for humans are minimized at a nearby point $b=.6$ (see the Supporting Information ~\ref{sec:optim} for per-game estimates).  ChatGPT-4's behavior exhibits the smallest error in that case, while the humans' behavior is the most varied, exhibits the highest errors, and is the least well-predicted of the three by $b=.5$.

The estimated $b$ varies across games, with the best fit being selfish ($b=1.0$) in the Ultimatum Game, but being centered around $b=.5$ in the other games (see Supporting Information Fig. ~\ref{fig:payoff_optim_MSE}).
We also perform a multinomial logistic discrete choice analysis and estimate the best fitting $b$'s by each game and find similar results (Supporting Information Table ~\ref{tab:beta-estimation}).

We also note that a linear specification does not fully capture preferences for relative payoffs as, for example, when $b=.5$ how a total payoff is allocated is inconsequential.   Instead, if one works with a CES (Constant Elasticity of Substitution) utility function \cite{mcfadden1963constant} of the form
$$\left(b \times (\text{Own Payoff})^{1/2} + (1-b) \times (\text{Partner Payoff})^{1/2}\right)^2,$$
then relative allocations across the two players are more distinguished.
For this specification we see the human error curve shift to have the weight that minimizes errors be more selfish, and we see more distinction between all three of ChatGPT-4, ChatGPT-3, and the humans.   This also carries over game by game as shown in the Supporting Information (see Fig. ~\ref{fig:payoff_optim_MSE_CES}).

\clearpage

\subsection*{Framing and Context}

Human behavior can be significantly altered by framing (e.g. \cite{tversky1989rational}).  We examine whether AI behavior also varies with how a given strategic setting is framed.
We find, that similar to humans, ChatGPT's decisions can be significantly influenced by changes in the framing or context of the same strategic setting. A request for an explanation of its decision, or asking them to act as if they come from some specific occupation can have an impact.

The Supporting Information 
has detailed prompts used for framing the AI (see Supporting Information ~\ref{app:steer}), and it also presents distributions of behaviors (Supporting Information ~\ref{sec:si-framing}). Here are some examples of how the framing matters.

When ChatGPT-3 is asked to explicitly explain its decision or when it is aware that the Dictator Game is witnessed by a third-party observer (a game host), it demonstrates significantly greater generosity as the dictator (see Supporting Information Fig. ~\ref{fig:steer-subfigA}).

In the Ultimatum Game, when ChatGPT-4 is made aware of the gender of the proposer (regardless of what it is), its decision as the responder moves away from the dominant strategy of accepting any proposal and starts demanding higher splits on average (Supporting Information Fig. ~\ref{fig:steer-subfigB}), even though we do not observe a specific gender effect.

In the Trust Game (see Fig. ~\ref{fig:steer-subfigD}, ~\ref{fig:steer-subfigE}, ~\ref{fig:steer-subfigF} in Supporting Information), as the size of the potential investment is increased, ChatGPT-4's strategy as the banker shifts from returning the original investment plus an even split of the profit (which equals a doubled investment) to evenly splitting the entire revenue (which is a tripled investment).  By contrast, ChatGPT-3 tends to make a more generous return to the investor when the potential investment is larger.

ChatGPT's decisions are also impacted when they are asked to play the games as if they are from a given occupation, altering their default role as a helpful assistant (see Supporting Information ~\ref{app:steer}). For instance, in the Ultimatum Game as the responder (Supporting Information Fig. ~\ref{fig:steer-subfigC}), when ChatGPT-4 is prompted to play as a mathematician, its decision shifts towards the dominant strategy, agreeing to accept as low as \$1 in most cases. Conversely, when prompted to be a legislator, its decisions align more with what is traditionally considered `fair': demanding \$50 in the majority of cases.

\subsection*{Learning}

One last thing we investigate is the extent to which the chatbots' behaviors change as they gain experience in different roles in a game, {\sl as if} they were learning from such experience.  This is something that is true of humans (e.g., \cite{benndorf2017experienced}).

In games with multiple roles (such as the Ultimatum Game and the Trust Game), the AIs' decisions can be influenced by previous exposure to another role. For instance, if ChatGPT-3 has previously acted as the responder in the Ultimatum Game, it tends to propose a higher offer when it later plays as the proposer, while ChatGPT-4's proposal remains unchanged (see Supporting Information Fig. ~\ref{fig:steer-order-subfigA}). Conversely, when ChatGPT-4 has previously been the proposer, it tends to request a smaller split as the responder (Fig. ~\ref{fig:steer-order-subfigB}).

Playing the banker's role in the Trust Game, especially when the investment is large, also influences ChatGPT-4 and ChatGPT-3's subsequent decisions as the investor, leading them to invest more (see Supporting Information Fig. ~\ref{fig:steer-order-subfigC}).
Similarly, having played the investor first also influences the AIs' subsequent decisions as the banker, resulting in both ChatGPT-3 and ChatGPT-4 returning more to the investor (see Supporting Information Fig. ~\ref{fig:steer-order-subfigD}).

Our analyses of learning and framing are far from systematic, and it would be interesting to compare how the effects of context change AI behavior to how context changes human behavior.  For example, it would be interesting to see how chatbots act when asked to assume the role of a specific gender, demographic group, or personality profile.


\section{Discussion}

We have sidestepped the question of whether artificial intelligence can think \cite{mitchell2023we,butlin2023consciousness,shapira2023clever}, which was a central point of Turing's original essay \cite{turing1950computing}, but we have performed a test along the lines of what he suggested.  We have found that AI and human behavior are remarkably similar.  Moreover, not only does AI's behavior sit within the human subject distribution in most games and questions, but it also exhibits signs of human-like complex behavior such as learning and changes in behavior from role-playing.
On the optimistic side, when AI deviates from human behavior, the deviations are in a positive direction: acting as if it is more altruistic and cooperative.
This may make AI well-suited for roles necessitating negotiation, dispute resolution, or caregiving, and may fulfill the dream of producing AI that is ``more human than human.'' This makes them potentially valuable in sectors such as conflict resolution, customer service, and healthcare.

The observation that ChatGPT's, especially ChatGPT-4's, behavior is more concentrated and consistent evokes both optimism and apprehension.  This is similar to what might happen if a single human were compared to the population.  However, the chatbots are used in technologies that interact with huge numbers of others and so this narrowness has consequences.   Positively, its rationality and constancy make AI highly attractive for various decision-making contexts and make it more stable and predictable. However, this also raises concerns regarding the potential loss of diversity in personalities and strategies (compared to the human population), especially when put into new settings and making important new decisions.

Our work establishes a straightforward yet effective framework and benchmark for evaluating chatbots and other AI as they are rapidly evolving. This may pave the way for a new field in AI behavioral assessment.
The AI that we tested was not necessarily programmed to pass this sort of Turing Test, and so that raises the question of when and how AI that is designed to converse with and inform humans, and is trained on human-generated data, necessarily behaves human-like more broadly.
That could also help in advancing our understanding of why humans exhibit certain traits.
Most importantly, the future will tell the extent to which AI enhances humans rather than substituting for them.\cite{rahwan2019machine,brynjolfsson2022turing}

In terms of limitations, given that our human data are collected from students, it is important to expand the reference population in further analyses.  The games we have chosen are prominent ones, but one can imagine expanding the suite of analyses included in a Turing Test, and also tailoring such tests to the specific tasks that are entrusted to different versions of AI.  In addition, the chatbots tested here are just one of a growing number, and a snapshot at a specific point in time of a rapidly evolving form of AI.  Thus, the results should not be taken as broadly representative, but instead should be taken as illustrative of a testing approach and what can be learned about particular instances of AI.

\bibliography{sn-bibliography}

\newpage

\appendix

\section*{Supporting Information}

\section{Method}

\subsection{Collecting Chatbot Responses}
\label{sec:SI-method-API}

We ran the interaction sessions with ChatGPT using the official API provided by OpenAI\footnote{OpenAI API: \url{https://platform.openai.com/}, retrieved 04/2023. }. 
In the main paper, ``ChatGPT-4'' refers to the model \texttt{gpt-4-0314} (snapshot of \texttt{gpt-4} from March 14th 2023), and ``ChatGPT-3'' refers to the model \texttt{gpt-3.5-turbo-0301} (snapshot of \texttt{gpt-3.5-turbo} from March 1st 2023)\footnote{OpenAI models: \url{https://platform.openai.com/docs/models}, retrieved 04/2023.}. 
We also tested the subscription-based ChatGPT Web version (Plus) and the freely available Web version (Free), retrieved in February 2023 through a third-party open source package \texttt{revChatGPT}\footnote{\texttt{revChatGPT}: \url{https://github.com/acheong08/ChatGPT}, retrieved 03/08/2023.}.

To avoid introducing confounders, we acquire the models' responses through the Chat Completion API\footnote{OpenAI Chat Completion API: \url{https://platform.openai.com/docs/api-reference/chat}, retrieved 03/2023. } with default parameters (sampling temperature as 1, number of chat completion choices as 1, and maximum number of tokens as infinity) and default \emph{system prompt} ``\texttt{You are a helpful assistant}.'' as suggested by the official guide\footnote{OpenAI Chat Completion API Guide: \url{https://platform.openai.com/docs/guides/gpt/chat-completions-api}, retrieved 03/2023. } unless otherwise specified (e.g., see Sections \ref{sec:big5} and \ref{app:steer}). 

Different snapshots of even the same chatbot version (e.g., ChatGPT-3), particularly the Web versions, can respond differently to the same query, and occasionally do not respond to queries. To mitigate this issue, it is advised to utilize the API versions or the OpenAI Platform Playground and specify the exact snapshot for testing a chatbot\footnote{The snapshots can be accessed and tested via both API and OpenAI Platform Playground (\emph{E.g.,} \url{https://platform.openai.com/playground?mode=chat&model=gpt-4-0314}, retrieved 12/28/2023).}. 
Valid actions rendered by the models are initially extracted from the AIs' responses using regular expressions or the \texttt{gpt-3.5-turbo} API. Subsequently, these extracted decisions undergo a manual verification process to ensure accuracy and relevance. Invalid responses are excluded from our analysis. More details can be found in the code and data repository: {\url{https://github.com/yutxie/ChatGPT-Behavioral}, retrieved 12/29/2023. 

\subsubsection{OCEAN Big Five Test}
\label{sec:big5}

The five-factor model (FFM), also known as the Big Five personality traits or the OCEAN model, was conceptualized in the 1980s and has been developed and refined over the past five decades. This model represents a comprehensive structure of traits that characterize individuals' personalities \cite{goldberg1993structure, mccrae1997personality, roccas2002big, mccrae2008five}. Each factor is defined by a cluster of intercorrelated traits known as facets. The five factors and their respective facets are, with descriptions following \cite{roccas2002big}:

\begin{itemize}
    \item \textbf{O}penness to experience: People with high scores in this trait are usually intellectual, imaginative, sensitive, and open-minded. Conversely, those with low scores are typically practical, less sensitive, and more traditional.
    \item \textbf{C}onscientiousness: Those who rank high in this dimension are generally careful, meticulous, responsible, organized, and principled. On the other hand, individuals with low scores in this trait often appear irresponsible, disorganized, and lacking in principles.
    \item \textbf{E}xtraversion: High scorers in Extraversion are often sociable, talkative, assertive, and energetic; whereas low scorers are more likely to be introverted, quiet, and cautious.
    \item \textbf{A}greeableness: Individuals scoring high on Agreeableness are often seen as amiable, accommodating, modest, gentle, and cooperative. In contrast, those with low scores may appear irritable, unsympathetic, distrustful, and rigid.
    \item \textbf{N}euroticism: People with high levels of Neuroticism tend to experience anxiety, depression, anger, and insecurity. Those scoring low on this trait are usually calm, composed, and emotionally stable.
\end{itemize}

In our study, we utilize the 50-item International Personality Item Pool (IPIP) representation of the Big Five factor structure, which is based on the markers developed by Goldberg \cite{goldberg1992development}. This questionnaire consists of 50 items in total, with each of the five factors represented by 10 questions. The IPIP representation provides a standardized and widely used measure for assessing individuals' personality traits within the framework of the Big Five model. The questionnaire is available in our released data and code repository. 

During the test phase, we adopt a specific approach for administering the questionnaire items to the chat models. For each item and chat model under investigation, we generate 30 independent chat instances. The system prompt is intentionally left empty to allow the chatbot models to generate responses based solely on the chat prompt provided.

The chat prompt for each questionnaire item follows a standardized format and includes the necessary instructions for participants. It is structured as follows: ``\texttt{The following item was rated on a five-point scale where 1=Disagree, 2=Partially Disagree, 3=Neutral, 4=Partially Agree, 5=Agree. Please select how the statement describes you and highlight your answer in [] (such as [1], [2], [3], [4], or [5]). + [Questionnaire item].}''

For example, a specific prompt might be: ``\texttt{The following item was rated on a five-point scale where 1=Disagree, 2=Partially Disagree, 3=Neutral, 4=Partially Agree, 5=Agree. Please select how the statement describes you and highlight your answer in [] (such as [1], [2], [3], [4], or [5]). I am the life of the party.}''

Note that some versions of ChatGPT are more likely to refuse to answer certain survey questions, resulting in more invalid responses for particular questions. In our experiments, the rate of invalid responses (averaged over questions) is 4.21\% (O: 2.79\%, C: 3.13\%, E: 4.44\%, A: 8.77\%, N: 1.92\%) for ChatGPT-4, and 11.17\% (O: 4.44\%, C: 29.81\%, E: 1.75\%, A: 17.35\%, N: 2.48\%) for ChatGPT-3. 
To accommodate the high invalid response rate for certain questions, we repeat each question multiple times (through independent API calls) until we obtain 30 valid responses.

\subsubsection{Behavioral Economics Games}

In behavioral economics games, we adopt a standardized procedure for generating instances and collecting responses from the chatbots. For each chatbot model, we generate 30 independent instances for each scenario by default, unless otherwise specified. These games are described within the chat prompts, accompanied by specific instructions, curated by two experts. The prompts of each game are available in our released code repository. 

For example, in the Dictator Game, the chat prompt used was as follows:
``\texttt{Hi, let's play a game. You are paired with another player. Your role is to decide how to divide \$100, and the other player simply receives your choice. How would you like to divide the money? Please give only one concrete choice and highlight the amount you give to the other player in [] (such as [\$x]).}''

The actions of a chatbot are extracted using regular expressions or the \texttt{gpt-3.5-turbo} API from its responses. To ensure accuracy and reliability, these responses undergo manual verification. 
As the invalid response rate is generally low for game queries in Dictator, Ultimatum, and Trust (1.33\% for ChatGPT-4 and 12.67\% for ChatGPT-3), we do not substitute invalid responses with additional API calls. 
For other games, we repeat each question multiple times (through independent API calls) until we obtain enough valid responses (30 for Public Goods, 80 for Bomb Risk, and 90 for Prisoner's Dilemma). 
Note that in the simulated games, the payoffs are fictitious and no real payments are made to the AI.

\subsubsection{Framing and Context}
\label{app:steer}

One objective of our research is to investigate the steerability of chatbot models. To accomplish this, we modify both the \emph{chat prompts} and \emph{system prompts} and assess whether the models' behaviors exhibit any noticeable changes. 

\paragraph{Witnesses and explanation requirement. }
\label{app:witness}

Our approach involves steering chatbot models through the imposition of witnesses and the requirement of explanations for their choices. 
Specifically, when addressing the Dictator Game scenario with a non-paired player, we introduce explicit information within the chat prompt, stating, ``\texttt{The game host hands you \$100. }'' 
Additionally, we prompt the models to provide explanations for their choices by incorporating the phrase, ``\texttt{Please explain your choice.}'' 

\paragraph{Gender. }

We further explore the influence of gender on the AIs' decision-making processes. To examine this, we specifically address the Dictator Game and the Ultimatum Game scenarios. In the chat prompt, we explicitly state, ``\texttt{You are paired with a male/female player.}'' 

\paragraph{Occupation specification. }

In our final investigation, we focus on studying the impact of specifying the occupation of chatbot models on their behaviors. To achieve this, we obtain occupation descriptions from the O*NET Database\footnote{O*NET Database: \url{https://www.onetcenter.org/database.html}, retrieved 03/2023. }. For each occupation, we include the occupation title, core tasks, and supplemental tasks within the system prompts. For example, 

\texttt{\\
You are a mathematician.\\
Your core tasks include:\\
* Address the relationships of quantities, ...\\
* Disseminate research by writing reports, ...\\
* Maintain knowledge in the field by ...\\
...\\
Your supplemental tasks include:\\
* Design, analyze, and decipher encryption systems ...\\}

The specific occupations we consider for our study encompass a range of professional roles, including mathematician, public relations specialist, journalist, investment fund manager, game theorist, teacher, and legislator. 
These occupations are particularly relevant as they have been identified as being highly exposed to large language models in a recent study conducted by OpenAI \cite{eloundou2023gpts}.

\subsection{Human Data}
\label{sec:SI-human-data}

\subsubsection{OCEAN Big Five Test}

\begin{figure}[htbp]
  \centering
  \begin{subfigure}[b]{0.45\textwidth}
    \centering
    \includegraphics[width=\textwidth]{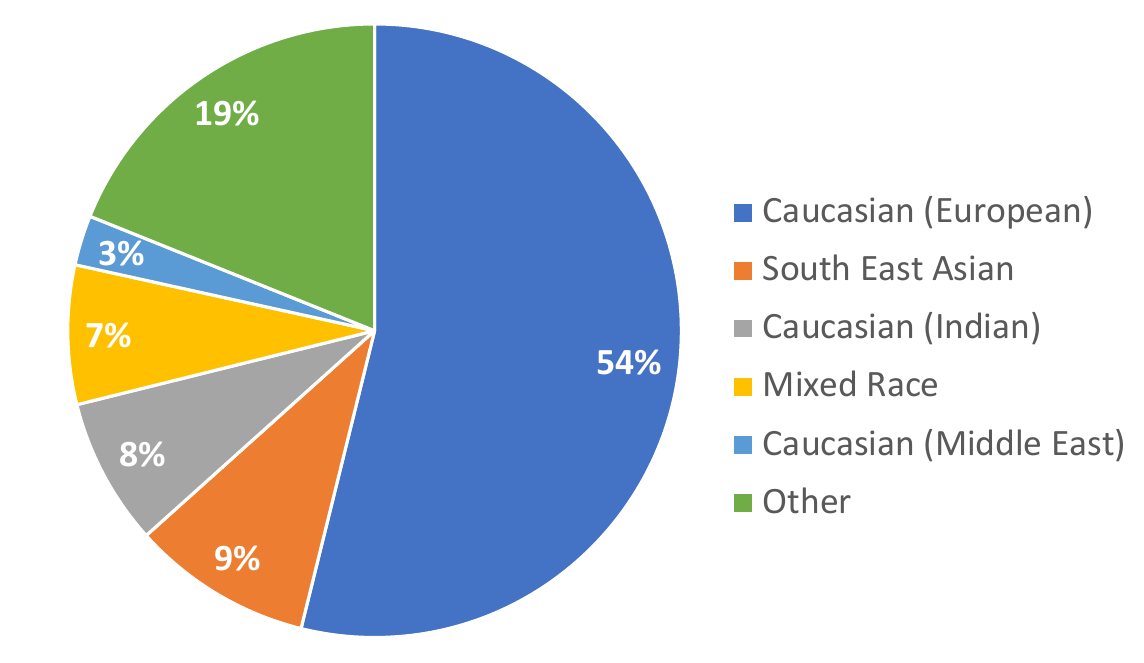}
    \caption{Respondent distribution over races. \\~\indent}
    \label{fig:demo-bigfive-race}
  \end{subfigure}%
  \hfill
  \begin{subfigure}[b]{0.50\textwidth}
    \centering
    \includegraphics[width=\textwidth]{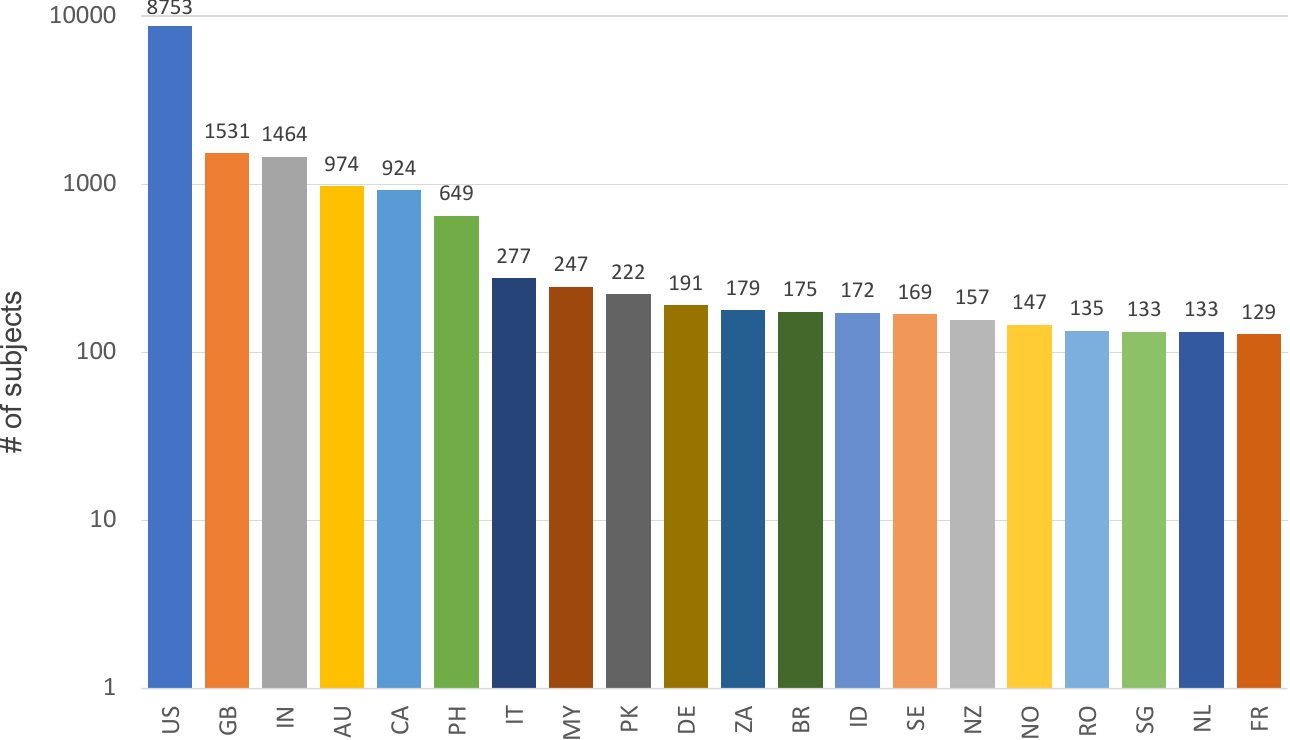}
    \caption{Respondent distribution over countries and regions. Only the top 20 are shown, covering 99.9\% of the data. }
    \label{fig:demo-bigfive-region}
  \end{subfigure}%
  \\
  \hspace{5pt}
  \begin{subfigure}[b]{0.33\textwidth}
    \begin{adjustbox}{left=2cm}
        \includegraphics[width=\textwidth]{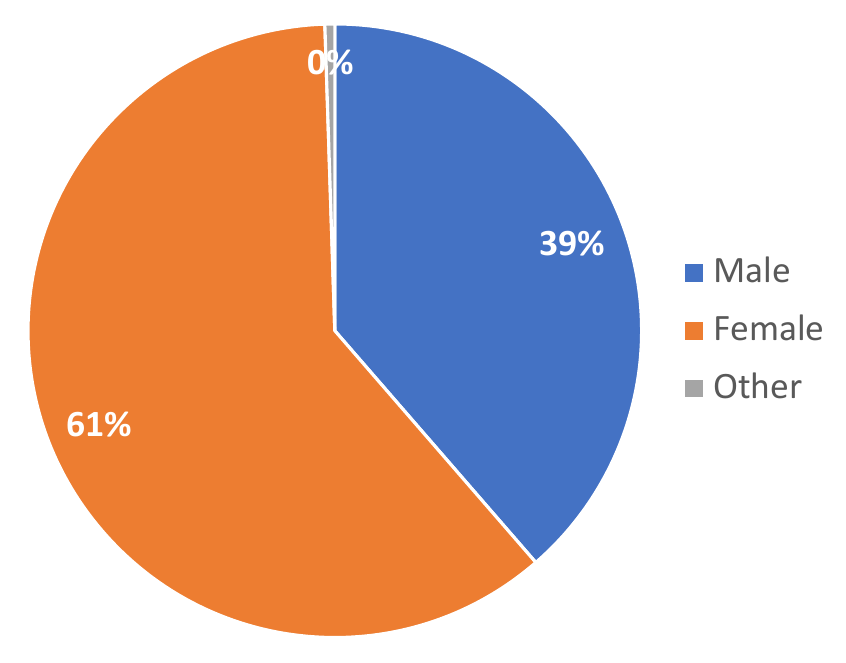}
    \end{adjustbox}
    \vspace{25pt}
    \caption{Respondent distribution over genders.}
    \label{fig:demo-bigfive-gender}
  \end{subfigure}%
  \hfill
  \begin{subfigure}[b]{0.45\textwidth}
    \centering
    \includegraphics[width=\textwidth]{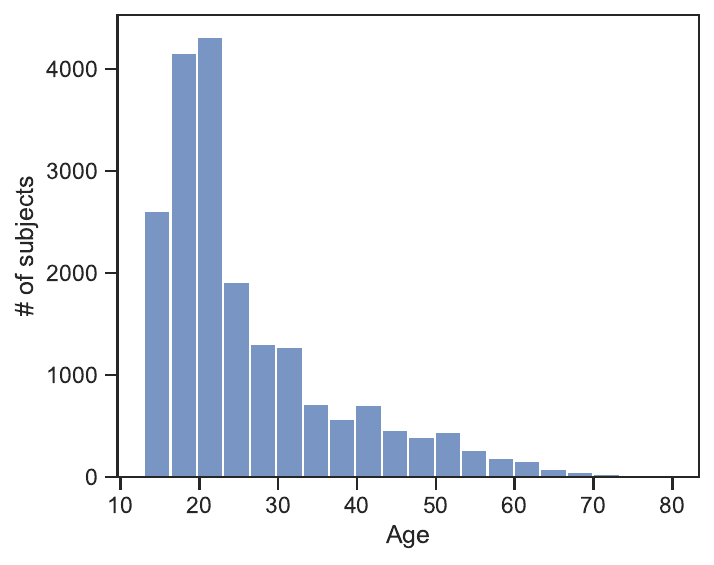}
    \caption{Respondent distribution over ages. Only ages under 80 are shown, covering 99.6\% of the data.}
    \label{fig:demo-bigfive-age}
  \end{subfigure}%
  \caption{Demographics of human respondents to the BigFive test. Category labels are extracted from the metadata in the codebook of the original dataset. }
  \label{fig:demo-bigfive}
\end{figure}

In our study, we utilize a publicly available database called the OCEAN Five Factor Personality Test Responses\footnote{OCEAN Five Factor Personality Test Responses dataset: \url{https://www.kaggle.com/datasets/lucasgreenwell/ocean-five-factor-personality-test-responses}, retrieved 03/2023. }. The data was sourced from the Open-Source Psychometrics Project \url{https://openpsychometrics.org/}, a nonprofit initiative aimed at both educating the public about psychology and collecting data for psychological research purposes. 

This database contains questions, answers, and metadata collected from a total of 19,719 tests. 
The subjects cover a wide range of demographic characteristics. The dataset comprises individuals from over 11 different races and 161 countries and regions, ensuring a diverse representation within the sample. Additionally, the age range of the participants spans from 13 years and above, encompassing a broad spectrum of age groups.
Regarding gender identification, participants self-identified as male, female, or other, acknowledging the importance of recognizing diverse gender identities.
Fig. \ref{fig:demo-bigfive} shows the distribution of demographics of the Big Five human responses data. 

\subsubsection{Behavioral Economics Games}

\begin{figure}[h]
  \centering
  \begin{subfigure}[b]{0.35\textwidth}
    \centering
    \includegraphics[width=\textwidth]{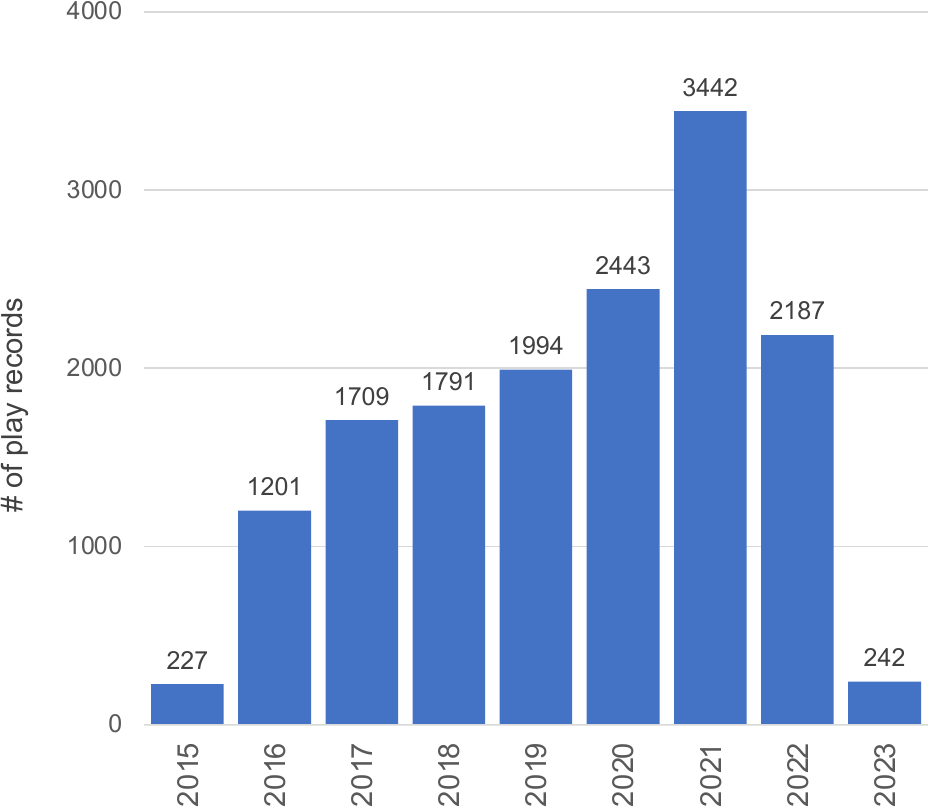}
    \vspace{20pt}
    \caption{Player record distribution over years. }
    \label{fig:demo-moblab-year}
  \end{subfigure}%
  \hfill
  \begin{subfigure}[b]{0.55\textwidth}
    \centering
    \includegraphics[width=\textwidth]{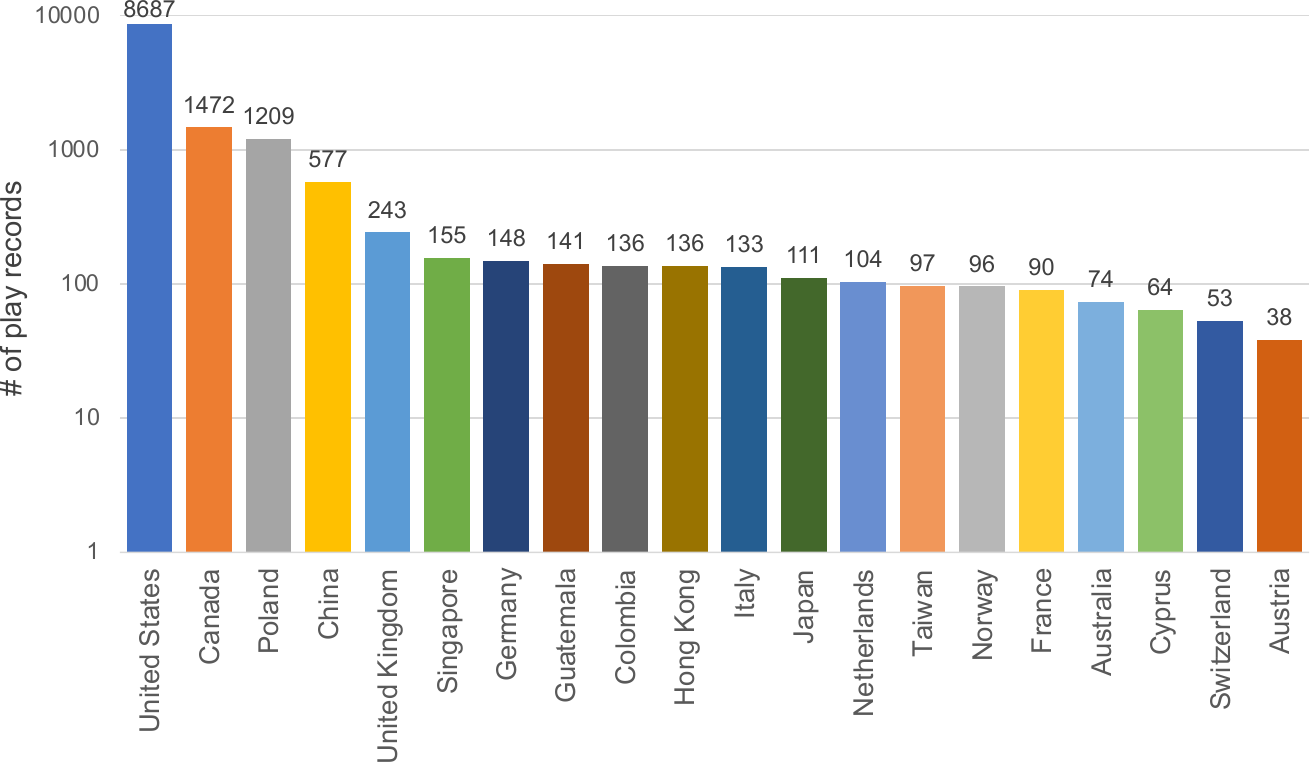}
    \caption{Player record distribution over countries and regions. Only the top 20 are shown, covering 97.9\% data. }
    \label{fig:demo-moblab-country}
  \end{subfigure}%
  \caption{Demographics of human players participating in MobLab behavioral economics games. }
  \label{fig:demo-moblab}
\end{figure}

The dataset under analysis comprises behavioral economic game data garnered via the MobLab Classroom platform\footnote{MobLab Classroom: \url{https://www.moblab.com/products/classroom}, retrieved 04/2023. } over an nine-year period from 2015 to 2023. This compendium of human behavioral data includes observations from 88,595 subjects, and 15,236 sessions, exhibiting a considerable geographical diversity, spanning 59 countries, and multiple continents. Participants are from nations and regions that encapsulate an array of socio-economic and cultural contexts, extending from the United States and Canada in North America, through Europe from Poland to the United Kingdom, and in Asia including China and Singapore. 
Fig. \ref{fig:demo-moblab} shows the distribution of demographics of the MobLab data in terms of game sessions.  Interested readers may refer to Lin et al. \cite{lin2020evidence} for more details about the demographics of the participants of MobLab games and variations across demographical groups.

\newpage
\section{Further Analysis}

\subsection{Payoffs}

ChatGPT's decisions are consistent with some forms of altruism, fairness, empathy, and reciprocity rather than maximization of its own payoff.  
To explore this in more detail, we calculate the payoff of the chatbots, the payoff of their (human) partner, and the combined payoff for both players in each game. These calculations are based on ChatGPT-4 and ChatGPT-3's strategies when paired with a player randomly drawn from the distribution of human players.
Similarly, we calculate the expected payoff of the human players when randomly paired with another human player.  
The results are presented in Table \ref{tab:payoff}.

\begin{table*}[htbp]
\centering
\caption{ChatGPT-4's strategies yield higher partner payoffs and higher combined payoffs compared to human players in all games where the payoff is not constant. ChatGPT-3 is the most cooperative among the three in the Public Goods game and the most altruistic in two games (Public Goods game and Trust game as the banker). Expected payoffs are calculated by sampling the partner's action from the human player distribution. The grey numbers after the ``$\pm$'' symbols are the standard errors based on 30 samples. 
}
\label{tab:payoff}
\resizebox{\columnwidth}{!}{%
\begin{tabular}{|c|c|c|c|c|}
\hline
\textbf{Game} & \textbf{Player} & \textbf{Selfish (own) payoff} & \textbf{Selfless (partner) payoff} & \textbf{Combined Payoff} \\ \hline
\multirow{3}{*}{Dictator} & Human & \$74.14\textcolor{gray}{\footnotesize $\ \pm\phantom{0}$0.19} & \$25.68\textcolor{gray}{\footnotesize $\ \pm\phantom{0}$0.19} & \$100.00\\
 & ChatGPT-4 & \$50.00\textcolor{gray}{\footnotesize $\ \pm$\phantom{0}0.00} & \$50.00\textcolor{gray}{\footnotesize $\ \pm$\phantom{0}0.00} & \$100.00 \\
 & ChatGPT-3 & \$64.83\textcolor{gray}{\footnotesize $\ \pm$\phantom{0}2.47} & \$35.17\textcolor{gray}{\footnotesize $\ \pm\phantom{0}$2.47} & \$100.00 \\ \hline
\multirow{3}{*}{{\begin{tabular}[c]{@{}c@{}}Ultimatum -\\ Proposer \end{tabular}}} & Human & \$33.51\textcolor{gray}{\footnotesize $\ \pm\phantom{0}$0.16} & \$35.19\textcolor{gray}{\footnotesize $\ \pm\phantom{0}$0.30} & \$68.70\textcolor{gray}{\footnotesize $\ \pm\phantom{0}$0.36}\\
 & ChatGPT-4 & \$45.98\textcolor{gray}{\footnotesize $\ \pm$\phantom{0}0.00} & \$45.98\textcolor{gray}{\footnotesize $\ \pm$\phantom{0}0.00} & \$91.96\textcolor{gray}{\footnotesize $\ \pm$\phantom{0}0.00} \\
 & ChatGPT-3 & \$35.10\textcolor{gray}{\footnotesize $\ \pm$\phantom{0}0.90} & \$32.79\textcolor{gray}{\footnotesize $\ \pm\phantom{0}$3.29} & \$67.89\textcolor{gray}{\footnotesize $\ \pm\phantom{0}$3.49} \\ \hline
\multirow{3}{*}{{\begin{tabular}[c]{@{}c@{}}Ultimatum -\\ Responder \end{tabular}}} & Human & \$35.19\textcolor{gray}{\footnotesize $\ \pm\phantom{0}$0.13} & \$33.51\textcolor{gray}{\footnotesize $\ \pm\phantom{0}$0.20} & \$68.70\textcolor{gray}{\footnotesize $\ \pm\phantom{0}$0.32} \\
 & ChatGPT-4 & \$37.60\textcolor{gray}{\footnotesize $\ \pm$\phantom{0}1.46} & \$39.60\textcolor{gray}{\footnotesize $\ \pm\phantom{0}$2.96} & \$77.20\textcolor{gray}{\footnotesize $\ \pm\phantom{0}$4.41} \\
 & ChatGPT-3 & \$38.26\textcolor{gray}{\footnotesize $\ \pm$\phantom{0}1.40} & \$36.75\textcolor{gray}{\footnotesize $\ \pm\phantom{0}$2.40} & \$75.00\textcolor{gray}{\footnotesize $\ \pm\phantom{0}$4.41} \\ \hline
\multirow{3}{*}{{\begin{tabular}[c]{@{}c@{}}Trust -\\ Investor \end{tabular}}} & Human & \$111.33\textcolor{gray}{\footnotesize $\ \pm\phantom{0}$0.11\phantom{0}} & \$76.03\textcolor{gray}{\footnotesize $\ \pm\phantom{0}$0.38} & \$187.36\textcolor{gray}{\footnotesize $\ \pm$\phantom{0}0.48\phantom{0}} \\
 & ChatGPT-4 & \$108.01\textcolor{gray}{\footnotesize $\ \pm$\phantom{0}0.48\phantom{0}} & \$84.13\textcolor{gray}{\footnotesize $\ \pm\phantom{0}$2.09} & \$192.14\textcolor{gray}{\footnotesize $\ \pm\phantom{0}$2.54\phantom{0}} \\
 & ChatGPT-3 & \$104.63\textcolor{gray}{\footnotesize $\ \pm$\phantom{0}0.84\phantom{0}} & \$68.04\textcolor{gray}{\footnotesize $\ \pm\phantom{0}$3.77} & \$172.67\textcolor{gray}{\footnotesize $\ \pm\phantom{0}$4.57\phantom{0}} \\ \hline
\multirow{3}{*}{{\begin{tabular}[c]{@{}c@{}}Trust -\\ Banker$^*$ \end{tabular}}} & Human & \$90.79\textcolor{gray}{\footnotesize $\ \pm\phantom{0}$0.97} & \$109.21\textcolor{gray}{\footnotesize $\ \pm\phantom{0}$0.97\phantom{0}} & \$200.00 \\
 & ChatGPT-4 & \$60.83\textcolor{gray}{\footnotesize $\ \pm\phantom{0}$2.26} & \$139.17\textcolor{gray}{\footnotesize $\ \pm\phantom{0}$2.26\phantom{0}} & \$200.00 \\
 & ChatGPT-3 & \$37.50\textcolor{gray}{\footnotesize $\ \pm\phantom{0}$5.99} & \$162.50\textcolor{gray}{\footnotesize $\ \pm\phantom{0}$5.99\phantom{0}} & \$200.00 \\ \hline
\multirow{3}{*}{Public Goods} & Human & \phantom{0}\$9.04\textcolor{gray}{\footnotesize $\ \pm$\phantom{0}0.02} & \phantom{0}\$9.04\textcolor{gray}{\footnotesize $\ \pm$\phantom{0}0.02} & \$36.15\textcolor{gray}{\footnotesize $\ \pm$\phantom{0}0.04} \\
 & ChatGPT-4 & \phantom{0}\$8.39\textcolor{gray}{\footnotesize $\ \pm$\phantom{0}0.05} & \phantom{0}\$9.69\textcolor{gray}{\footnotesize $\ \pm$\phantom{0}0.05} & \$37.45\textcolor{gray}{\footnotesize $\ \pm$\phantom{0}0.10} \\
 & ChatGPT-3 & \phantom{0}\$7.97\textcolor{gray}{\footnotesize $\ \pm$\phantom{0}0.13} & \$10.10\textcolor{gray}{\footnotesize $\ \pm$\phantom{0}0.13} & \$38.28\textcolor{gray}{\footnotesize $\ \pm$\phantom{0}0.27} \\ \hline
\multirow{3}{*}{{\begin{tabular}[c]{@{}c@{}}Prisoner's \\Dilemma$^\dagger$ \end{tabular}}} & Human & \$345.12\textcolor{gray}{\footnotesize $\ \pm\phantom{0}$0.53\phantom{0}} & \$345.12\textcolor{gray}{\footnotesize $\ \pm$\phantom{0}0.70\phantom{0}} & \$690.24\textcolor{gray}{\footnotesize $\ \pm$\phantom{0}0.18\phantom{0}} \\
 & ChatGPT-4 & \$205.48\textcolor{gray}{\footnotesize $\ \pm$10.70\phantom{0}} & \$531.31\textcolor{gray}{\footnotesize $\ \pm$14.27\phantom{0}} & \$736.79\textcolor{gray}{\footnotesize $\ \pm$\phantom{0}3.57\phantom{0}} \\
 & ChatGPT-3 & \$250.48\textcolor{gray}{\footnotesize $\ \pm$16.38\phantom{0}} & \$471.31\textcolor{gray}{\footnotesize $\ \pm$21.84\phantom{0}} & \$721.79\textcolor{gray}{\footnotesize $\ \pm$\phantom{0}5.46\phantom{0}} \\ \hline
\end{tabular}
}
\vspace{5pt}

$^*:$ To be comparable, the Trust-Banker calculations are done assuming that the original investment is \$50.  \\ $^\dagger:$ The Prisoner's Dilemma reports the payoffs in the first round of the game.
\end{table*}

Table~\ref{tab:payoff} shows that ChatGPT-4 outperforms human players in terms of expected own payoff only in the Ultimatum Game, and a lower own payoff in the other games involving trust or cooperation. However, in all seven game scenarios, it obtains a higher expected payoff for its partner. Moreover, it achieves a higher combined payoff for both players in five out of seven games, the exception being the Dictator game and the Trust game as the banker (where the combined payoff is constant). 

These findings are consistent with an increased level of altruism and cooperation compared to the human player distribution. On the other hand, ChatGPT-3 obtains payoffs that are closer to humans in the Dictator game, the Ultimatum Game as the proposer, and Prisoner's Dilemma. 
And, although it yields a lower own payoff in the Trust games and the Public Goods game compared to ChatGPT-4, it achieves an even higher partner payoff and combined payoff in the Public Goods game and a higher partner payoff as the banker in the Trust game. 

\subsection{Optimization Objective}
\label{sec:optim}

The findings presented above indicate that the strategic outputs of ChatGPT-4 and ChatGPT-3 models tend to yield higher partner payoffs compared to human players, with ChatGPT-4 frequently attaining the highest combined payoff. This subsection is dedicated to a systematic exploration of the preferences that best predict behavior.

Discerning the intrinsic objectives of models can be challenging when solely examining their training methodologies. Taking ChatGPT-4 \citep{openai2023gpt4} as an example, which serves as the backbone of ChatGPT-4, it is a Transformer-style model initially pre-trained to forecast the subsequent token in a given document. This pre-training phase leverages publicly accessible data such as internet-sourced information. In this stage, the primary objective function is essentially to maximize the likelihood of a word’s occurrence when provided with the preceding words for each sentence or document within the training data. Subsequently, the model undergoes a fine-tuning process using Reinforcement Learning from Human Feedback (RLHF) \citep{ouyang2022training}. The fine-tuning tasks employed encompass natural language processing activities such as text generation, question answering, dialog generation, summarization, and information extraction. The process involves presenting demonstration responses and having human evaluators rank the outputs from best to worst.  With RLHF, OpenAI may also add restrictions regarding safety considerations\footnote{OpenAI usage policies: \url{https://openai.com/policies/usage-policies}, retrieved 03/2023.}.  Notably, these tasks and the safety policies do not inherently align with decision-making tasks, and this approach does not directly translate into a well-defined objective function for behavioral games.  To the best of our knowledge, there is no evidence that either ChatGPT-3 or ChatGPT-4 was fine-tuned to behave in specific directions in the tests included in our study.  Hence, we rely on observations of models’ behaviors to understand their objective.

In the context of behavioral games, where the player is presumed to optimize a blend of selfish and partner payoffs, the optimization objective function can be formulated as a constant elasticity of substitution (CES) utility function \cite{mcfadden1963constant}:
\begin{align}
    \label{eq:interpolation}
    U_b &= \left[b\cdot S^r+(1-b)\cdot P^r \right]^{(1/r)},
\end{align}
where $U_b$ denots the utility function, $S$ is the player's selfish (own) payoff, $P$ is the selfless (partner) payoff, $b\in[0,1]$ is the weight, and $r$ is the CES parameter. 

The selfish and selfless expected payoff curves are plotted in Fig. \ref{fig:payoff-curves}, which also contains examples of the expected utility function when $b=1/2$ for two CES specifications (the linear specification with $r=1$ and the non-linear specification with $r=1/2$).
These are the expected utilities against the distribution of play from the human partner distribution, as a function of the strategy played.

\begin{figure}[htbp]
    \centering
    \begin{subfigure}[b]{0.30\textwidth}
        \centering
        \includegraphics[width=\textwidth]{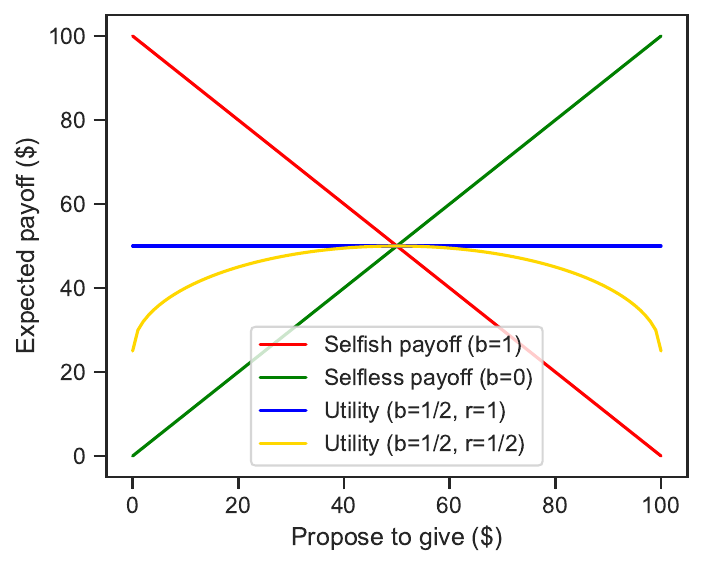}
        \caption{Dictator}
        \label{fig:payoff-curve-dictator}
    \end{subfigure}%
    \hfill
    \begin{subfigure}[b]{0.30\textwidth}
        \centering
        \includegraphics[width=\textwidth]{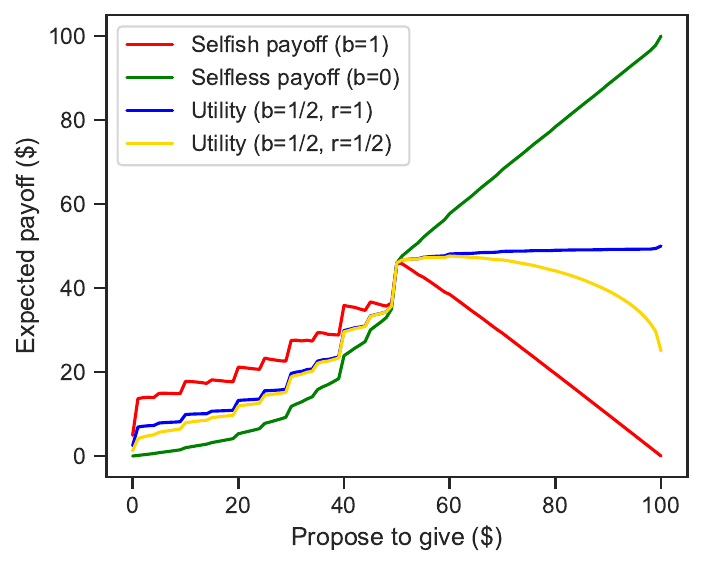}
        \caption{Ultimatum - Proposer}
        \label{fig:payoff-curve-ultimatum-1}
    \end{subfigure}%
    \hfill
    \begin{subfigure}[b]{0.30\textwidth}
        \centering
        \includegraphics[width=\textwidth]{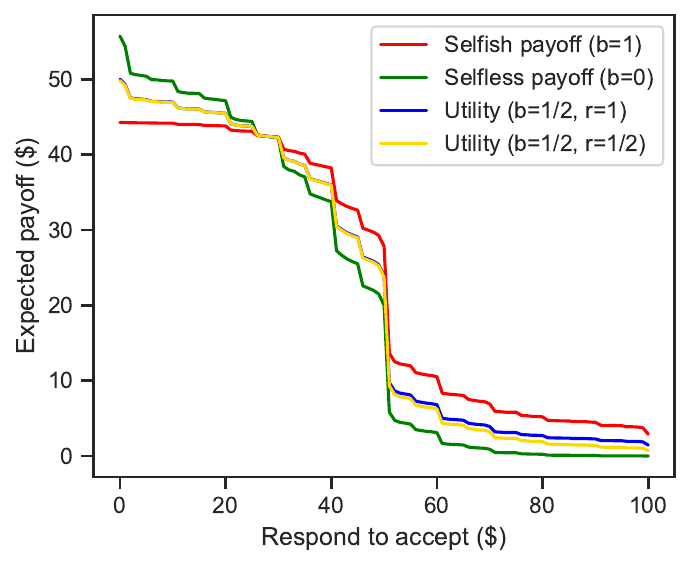}
        \caption{Ultimatum - Responder}
        \label{fig:payoff-curve-ultimatum-2}
    \end{subfigure}%
    \vspace{15pt}
    \begin{subfigure}[b]{0.30\textwidth}
        \centering
        \includegraphics[width=\textwidth]{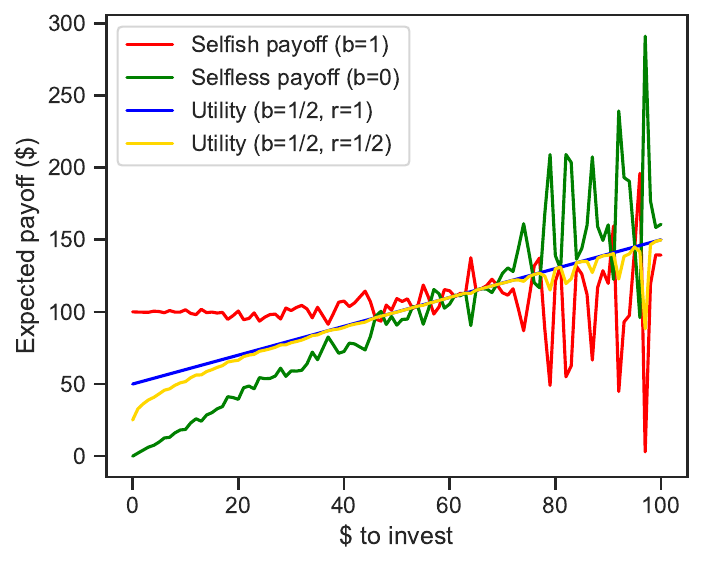}
        \caption{Trust - Investor}
        \label{fig:payoff-curve-trust-1}
    \end{subfigure}%
    \hfill
    \begin{subfigure}[b]{0.30\textwidth}
        \centering
        \includegraphics[width=\textwidth]{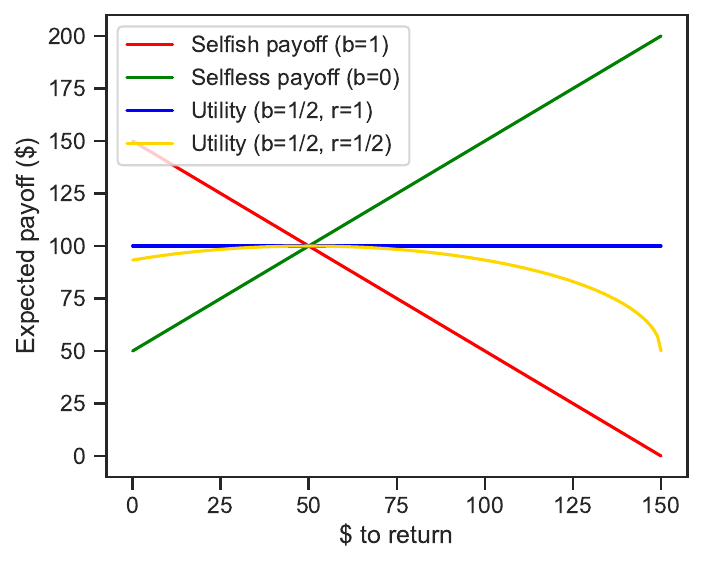}
        \caption{Trust - Banker (with \$50 invested)}
        \label{fig:payoff-curve-trust-3}
    \end{subfigure}%
    \hfill
    \begin{subfigure}[b]{0.30\textwidth}
        \centering
        \includegraphics[width=\textwidth]{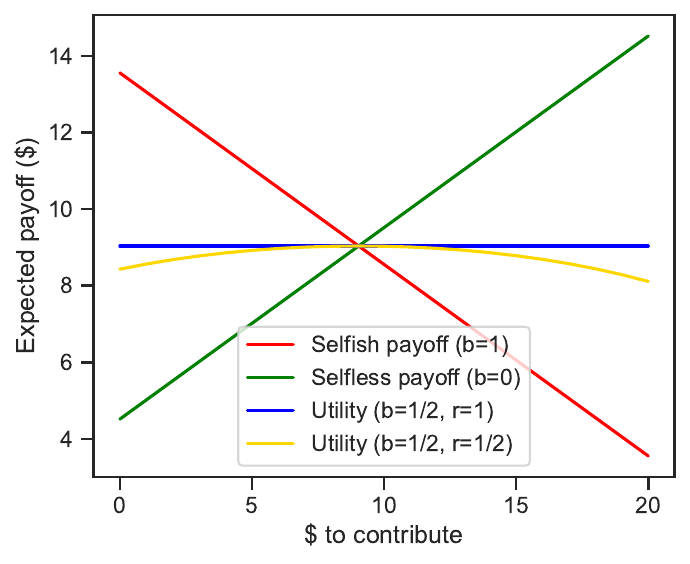}
        \caption{Public Goods}
        \label{fig:payoff-curve-PG}
    \end{subfigure}%
  
    \caption{
    Expected selfish (own) payoff (red lines) and selfless (partner) payoff (green lines) of every single action with a randomly sampled human partner.
    Blue and yellow lines show the weighted expected utility function examples as defined in Eq. \ref{eq:interpolation}. When $r=1$ and $b=1/2$ (blue lines), the utility function becomes the overall welfare (average payoff). 
    Weighted utility functions for other values of $b$ can be obtained similarly. 
    }
\label{fig:payoff-curves}
\end{figure}

For every game and parameter value $b$ ($0 \leq b \leq 1$) for the utility function, we can calculate the mean square error (MSE) for any given action compared to the best response as 
\begin{equation}
    \text{MSE}_b = \frac{1}{\vert\mathcal{O}\vert}
    \sum_{k\in\mathcal{O}}\left[
    1-\frac{U_b(k)}{U_b^*}
    \right]^2,
\end{equation}
where $\mathcal{O}$ is the set of observations, $k$ is an action choice from the observation (e.g., give \$50 in the Dictator game), $U_b(k)$ is the expected utility from action $k$ calculated with the expected selfish payoff $S(k)$ and the partner payoff $P(k)$, and $U_b^*$ is the theoretical maximum utility from the best response (i.e., based on Fig. \ref{fig:payoff-curves}). 

The results are shown in Fig. \ref{fig:payoff_optim_MSE} (linear CES utility specification with $r=1$) and Fig. \ref{fig:payoff_optim_MSE_CES} (non-linear CES specification with $r=1/2$).

For the linear specification and most of the games except for Ultimatum, human players and ChatGPTs achieve their lowest optimization MSE at around $b=1/2$. We also note that ChatGPTs tend to have smaller optimization errors compared to humans when $b \le 0.5$, showing higher optimization efficiency when the objective is less selfish.

\begin{figure*}[htbp]
  \centering
  \begin{subfigure}[b]{0.24\textwidth}
    \centering
    \includegraphics[width=\textwidth]{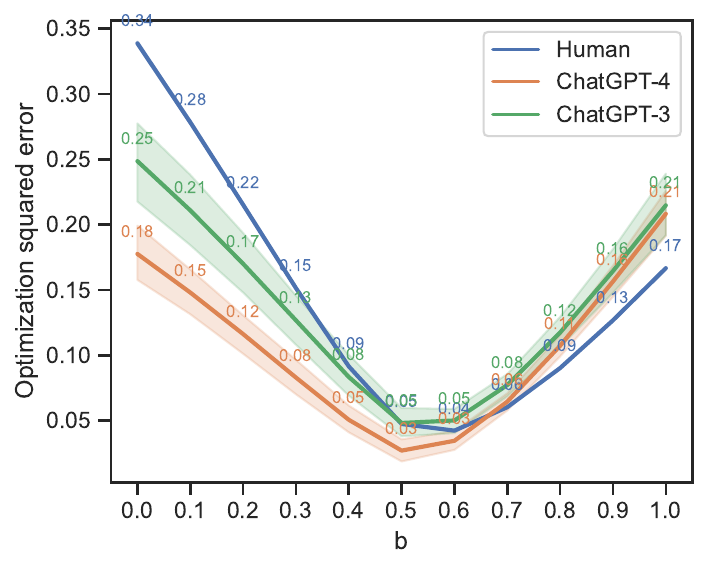}
    \caption{Average \ \\}
    \label{fig:MSE_avg}
  \end{subfigure}%
  \hfill
  \begin{subfigure}[b]{0.24\textwidth}
    \centering
    \includegraphics[width=\textwidth]{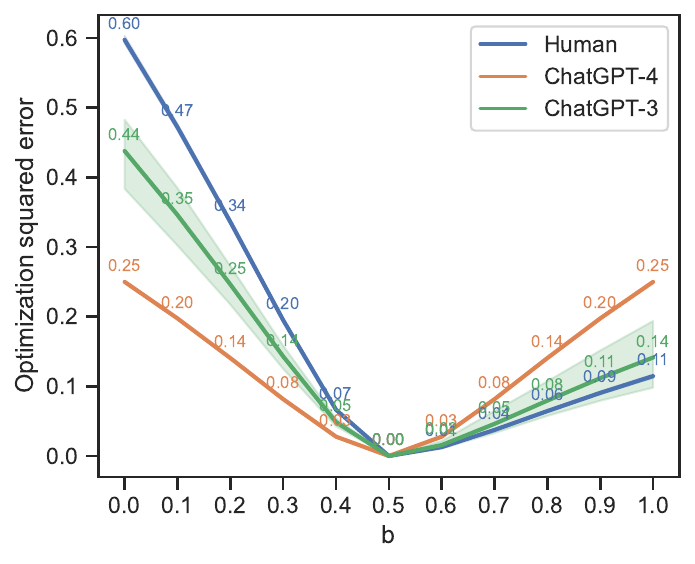}
    \caption{Dictator \ \\}
    \label{fig:MSE_dictator}
  \end{subfigure}%
  \hfill
  \begin{subfigure}[b]{0.24\textwidth}
    \centering
    \includegraphics[width=\textwidth]{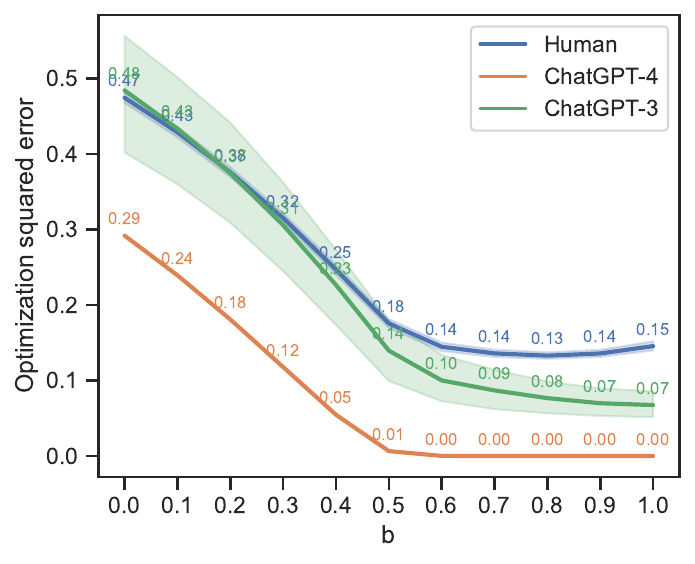}
    \caption{Ultimatum - Proposer}
    \label{fig:MSE_ultimatum_1}
  \end{subfigure}%
  \hfill
  \begin{subfigure}[b]{0.24\textwidth}
    \centering
    \includegraphics[width=\textwidth]{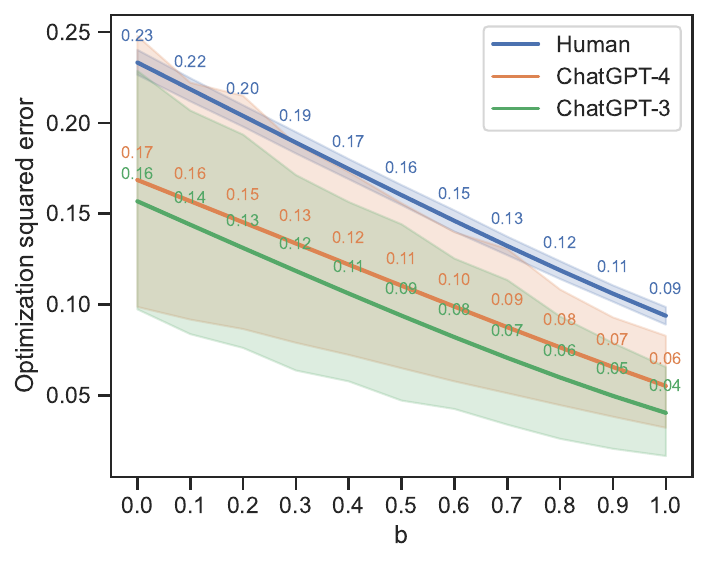}
    \caption{Ultimatum - Responder}
    \label{fig:MSE_ultimatum_2}
  \end{subfigure}%
  
  \vspace{0.5cm}
  
  \begin{subfigure}[b]{0.24\textwidth}
    \centering
    \includegraphics[width=\textwidth]{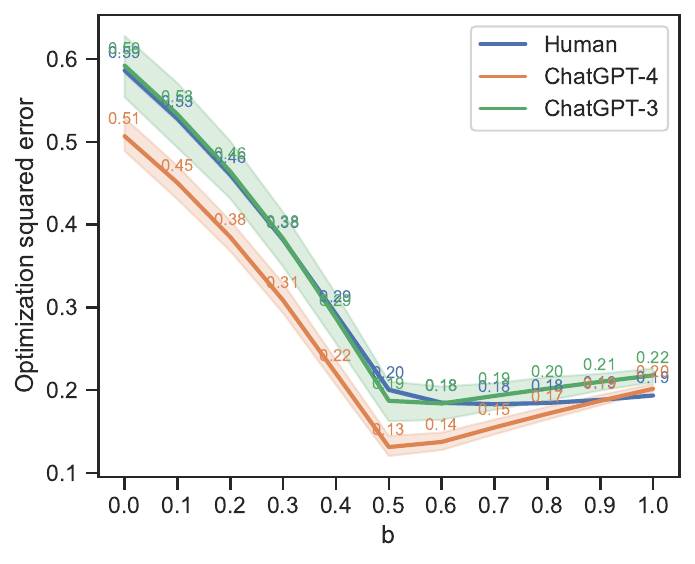}
    \caption{Trust - Investor \ \\}
    \label{fig:MSE_trust_1}
  \end{subfigure}%
  \hfill
  \begin{subfigure}[b]{0.24\textwidth}
    \centering
    \includegraphics[width=\textwidth]{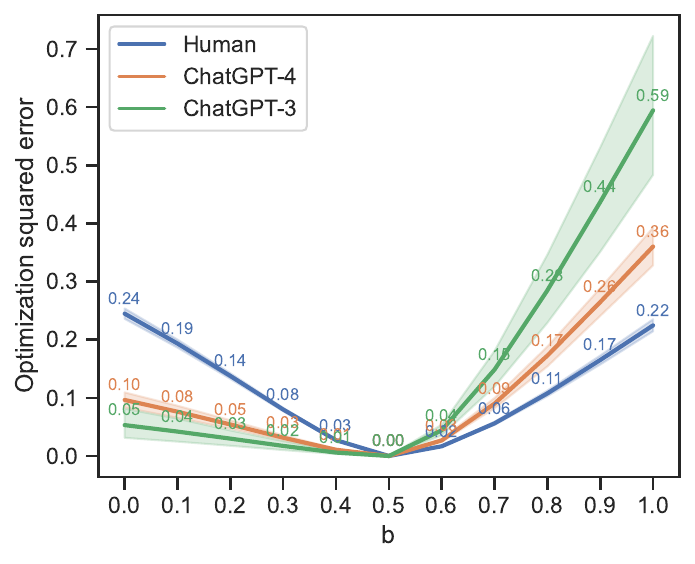}
    \caption{Trust - Banker \ \\}
    \label{fig:subfigF}
  \end{subfigure}%
  \hfill
  \begin{subfigure}[b]{0.24\textwidth}
    \centering
    \includegraphics[width=\textwidth]{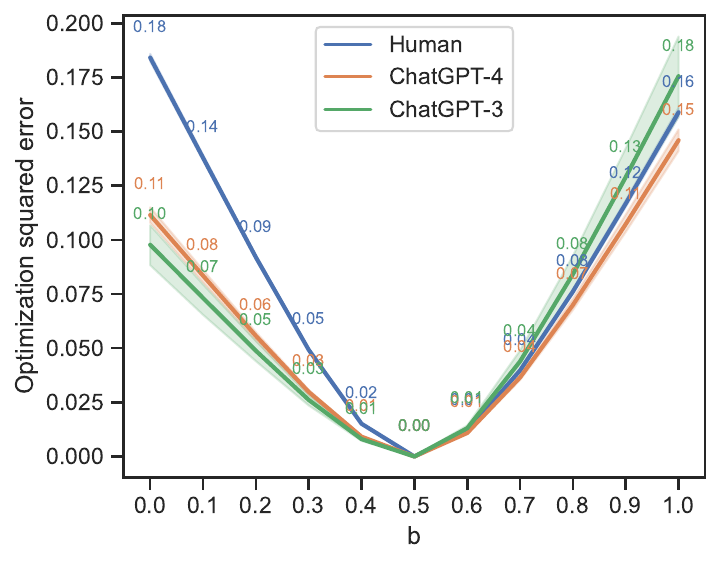}
    \caption{Public Goods \ \\}
    \label{fig:MSE_PG}
  \end{subfigure}%
  \hfill
  \begin{subfigure}[b]{0.24\textwidth}
    \centering
    \includegraphics[width=\textwidth]{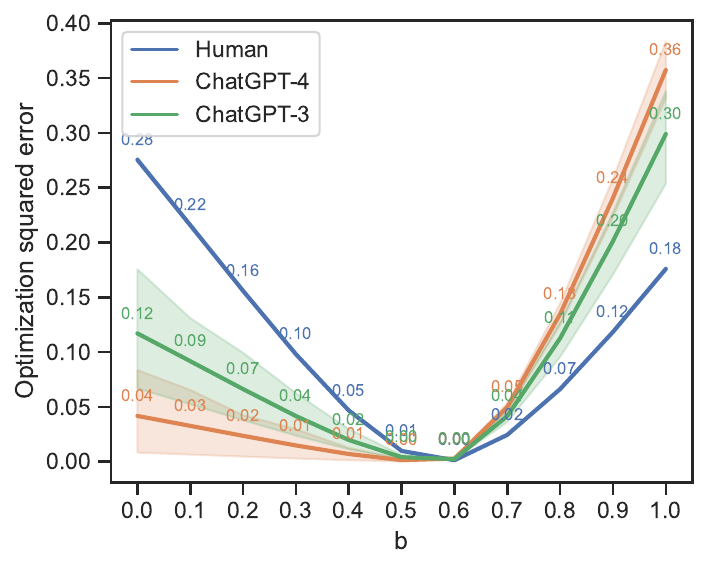}
    \caption{Prisoner's Dilemma}
    \label{fig:MSE_PD}
  \end{subfigure}%
  \caption{Mean squared error of the actual distribution of play relative to the best-response payoff, when matched with a partner playing the human distribution. The average is across all games. The errors are reported for each possible $b$, which is the weight on own vs partner payoff in the utility function (linear blend, with CES specification $r=1$). $b = 1$ is the purely selfish (own) payoff, $b = 0$ is the purely selfless (partner) payoff, and $b = 1/2$ is the overall welfare (average) payoff. The values of mean square errors are annotated in the plots. 
  }
  \label{fig:payoff_optim_MSE}
\end{figure*}

\begin{figure*}[htbp]
  \centering
  \begin{subfigure}[b]{0.24\textwidth}
    \centering
    \includegraphics[width=\textwidth]{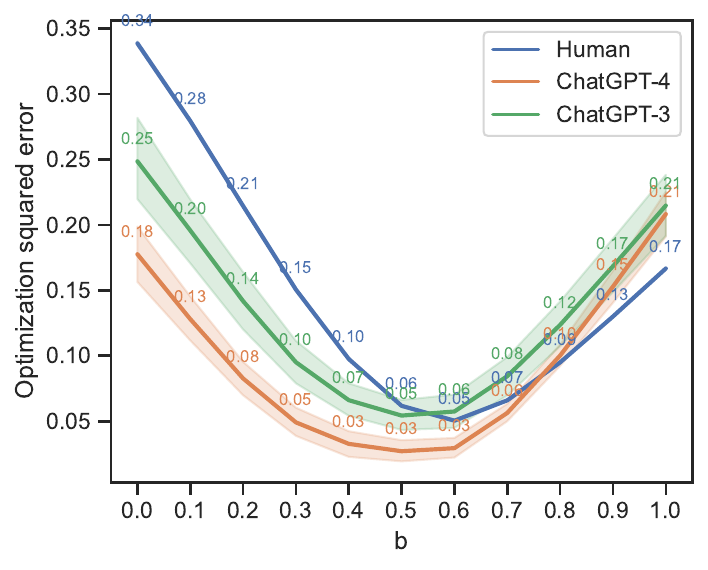}
    \caption{Average \ \\}
  \end{subfigure}%
  \hfill
  \begin{subfigure}[b]{0.24\textwidth}
    \centering
    \includegraphics[width=\textwidth]{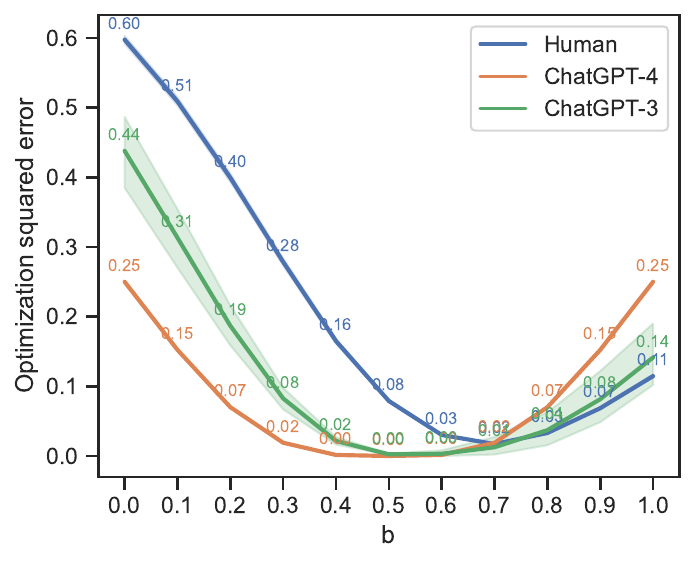}
    \caption{Dictator \ \\}
  \end{subfigure}%
  \hfill
  \begin{subfigure}[b]{0.24\textwidth}
    \centering
    \includegraphics[width=\textwidth]{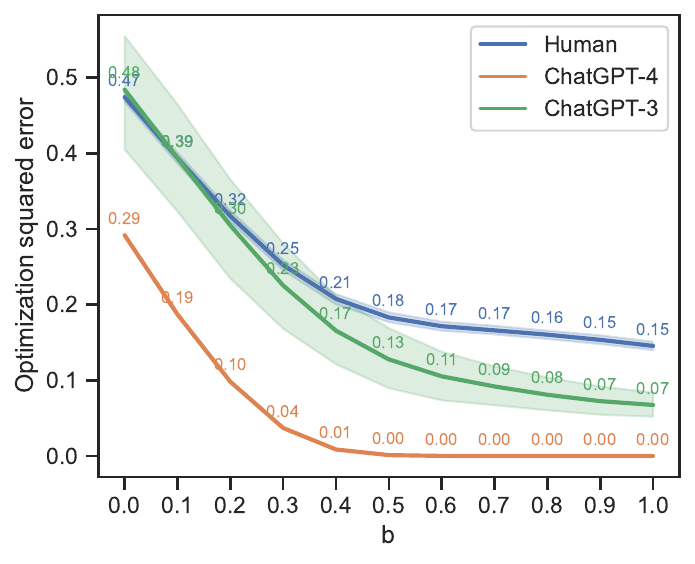}
    \caption{Ultimatum - Proposer}
  \end{subfigure}%
  \hfill
  \begin{subfigure}[b]{0.24\textwidth}
    \centering
    \includegraphics[width=\textwidth]{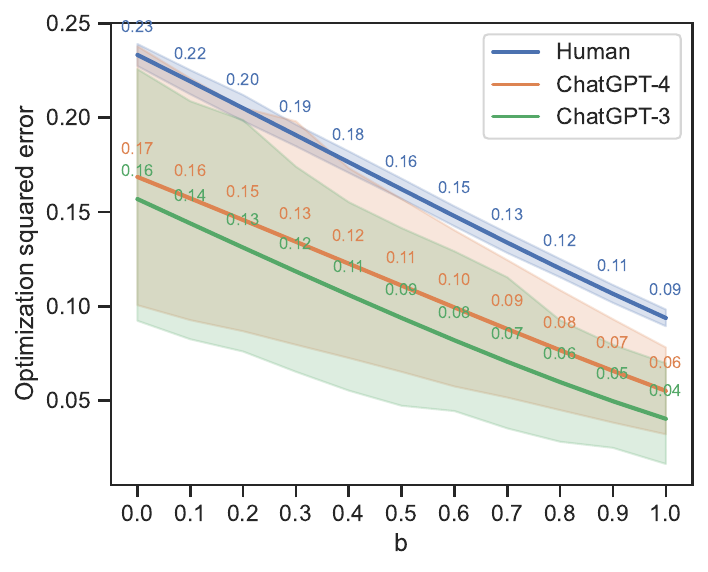}
    \caption{Ultimatum - Responder}
  \end{subfigure}%
  
  \vspace{0.5cm}
  
  \begin{subfigure}[b]{0.24\textwidth}
    \centering
    \includegraphics[width=\textwidth]{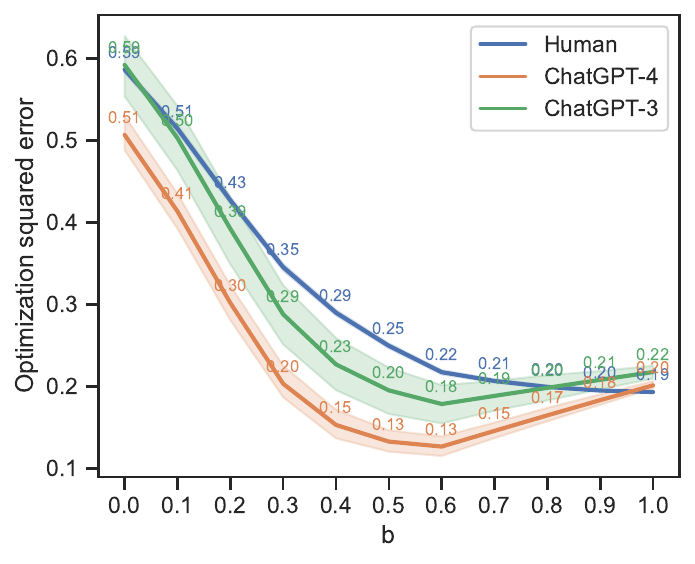}
    \caption{Trust - Investor \ \\}
  \end{subfigure}%
  \hfill
  \begin{subfigure}[b]{0.24\textwidth}
    \centering
    \includegraphics[width=\textwidth]{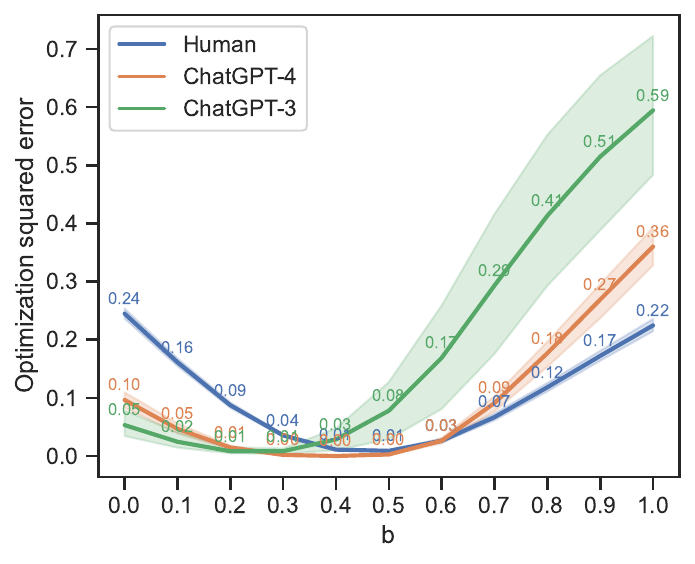}
    \caption{Trust - Banker \ \\}
  \end{subfigure}%
  \hfill
  \begin{subfigure}[b]{0.24\textwidth}
    \centering
    \includegraphics[width=\textwidth]{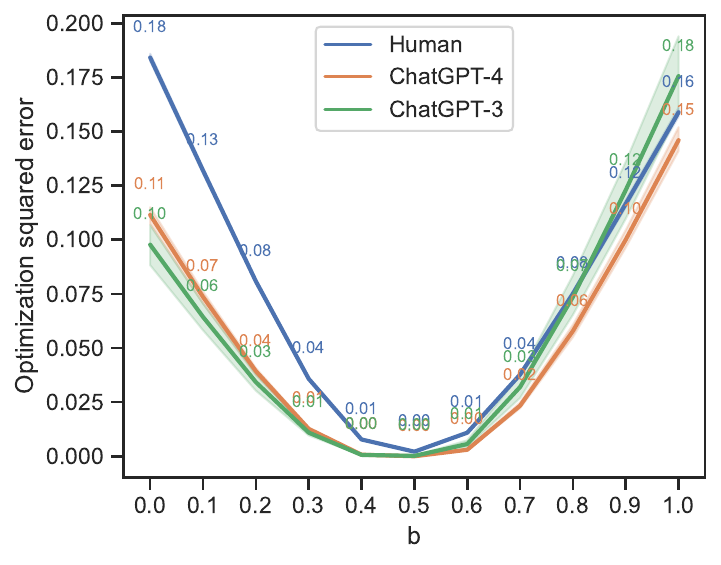}
    \caption{Public Goods \ \\}
  \end{subfigure}%
  \hfill
  \begin{subfigure}[b]{0.24\textwidth}
    \centering
    \includegraphics[width=\textwidth]{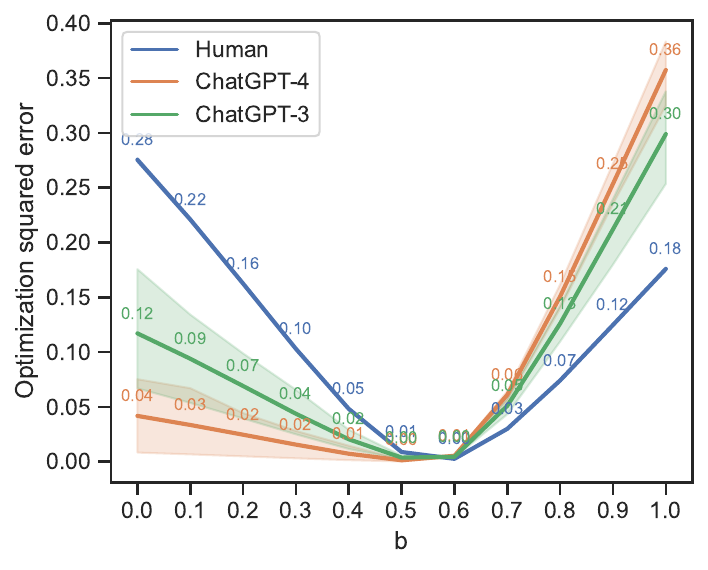}
    \caption{Prisoner's Dilemma}
  \end{subfigure}%
  \caption{Mean squared error of the actual distribution of play relative to the best-response payoff, when matched with a partner playing the human distribution. The average is across all games. The errors are reported for each possible $b$, which is the weight on own vs partner payoff in the utility function (non-linear blend, with CES specification $r=1/2$). $b = 1$ is the purely selfish (own) payoff, and $b = 0$ is the purely selfless (partner) payoff. The values of mean square errors are annotated in the plots. 
  }
  \label{fig:payoff_optim_MSE_CES}
\end{figure*}

Beyond the revealed preference analysis above,  
we also estimate the parameter $b$ for each game and model, positing a logistic multinomial model framework (as in McFadden's discrete choice problem \cite{mcfadden1973conditional}).
We only do this for the linear utility specification, as that is the usual multinomial logit formulation.

According to this framework, actions are sampled in accordance with the following probability distribution:
\begin{equation}
    \text{Pr}(k) = \frac{\exp(U_b(k))}{\sum\limits_{j\le K}\exp(U_b(j))},
\end{equation}
where $K$ is the number of possible action choices. 

The estimation results are presented in Table \ref{tab:beta-estimation}, which are well aligned with those in Fig. \ref{fig:payoff_optim_MSE}. For many games, including Dictator, Trust - Banker, Public Goods (for only ChatGPT-3), and Prisoner's Dilemma, the estimated $b$ values from ChatGPTs are significantly smaller than humans, indicating ChatGPTs behave as if they were less selfish behavioral than humans.

\begin{table*}[htbp]
\centering
\caption{Estimation of the weight $b$ by multinomial logit discrete choice analysis. Green highlights the cases when the estimated $b$ for ChatGPT models is significantly smaller (less weight on own payoff) than the estimate for humans.
}
\label{tab:beta-estimation}
\resizebox{\columnwidth}{!}{%
\begin{tabular}{|c|c|c|c|c|c|c|c|}
\hline
\multirow{2}{*}{\textbf{Game}} & \multirow{2}{*}{\textbf{Player}} & \multicolumn{3}{c|}{\textbf{With CES specification $r=1$}} & \multicolumn{3}{c|}{\textbf{With CES specification $r=1/2$}} \\ \hhline{~~------}
& & \textbf{Estimated $b$} & \textbf{Standard error} & \textbf{Confidence interval} & \textbf{Estimated $b$} & \textbf{Standard error} & \textbf{Confidence interval} \\ \hline
\multirow{3}{*}{Dictator} & Human & 0.517 & 0.000 & (0.516, 0.517) & 0.658 & 0.000 & (0.657, 0.659) \\
 & ChatGPT-4 & \colorbox{cellgreen}{0.500} & 0.003 & (0.494, 0.506) & \colorbox{cellgreen}{0.500} & 0.009 & (0.482, 0.518) \\
 & ChatGPT-3 & \colorbox{cellgreen}{0.509} & 0.003 & (0.502, 0.516) & \colorbox{cellgreen}{0.582} & 0.010 & (0.563, 0.601) \\ \hline
\multirow{3}{*}{{\begin{tabular}[c]{@{}c@{}}Ultimatum -\\ Proposer \end{tabular}}} & Human & 1.000 & 0.005 & (0.989, 1.011) & 1.000 & 0.005 & (0.990, 1.01) \\
& ChatGPT-4 & 1.000 & 0.076 & (0.851, 1.149) & 1.000 & 0.079 & (0.845, 1.155) \\
& ChatGPT-3 & 1.000 & 0.076 & (0.851, 1.149) & 1.000 & 0.211 & (0.587, 1.413) \\ \hline
\multirow{3}{*}{{\begin{tabular}[c]{@{}c@{}}Ultimatum -\\ Responder \end{tabular}}} & Human & 1.000 & 0.005 & (0.990, 1.010) & 1.000 & 0.006 & (0.989, 1.011) \\
& ChatGPT-4 & 1.000 & 0.070 & (0.862, 1.138) & 1.000 & 0.079 & (0.845, 1.155) \\
& ChatGPT-3 & 1.000 & 0.070 & (0.862, 1.138) & 1.000 & 0.077 & (0.849, 1.151) \\ \hline
\multirow{3}{*}{{\begin{tabular}[c]{@{}c@{}}Trust -\\ Investor \end{tabular}}} & Human & 0.535 & 0.000 & (0.535, 0.535) & 0.570 & 0.000 & (0.569, 0.570) \\
& ChatGPT-4 & 0.532 & 0.003 & (0.526, 0.538) & 0.566 & 0.003 & (0.559, 0.572) \\
& ChatGPT-3 & 0.535 & 0.003 & (0.529, 0.541) & 0.569 & 0.003 & (0.564, 0.575) \\ \hline
\multirow{3}{*}{{\begin{tabular}[c]{@{}c@{}}Trust -\\ Banker$^*$ \end{tabular}}} & Human & 0.504 & 0.000 & (0.504, 0.505) & 0.475 & 0.001 & (0.473, 0.477) \\
& ChatGPT-4 & \colorbox{cellgreen}{0.496} & 0.002 & (0.492, 0.500) & \colorbox{cellgreen}{0.395} & 0.007 & (0.382, 0.408) \\
& ChatGPT-3 & \colorbox{cellgreen}{0.488} & 0.003 & (0.482, 0.495) & \colorbox{cellgreen}{0.300} & 0.009 & (0.283, 0.318) \\ \hline
\multirow{3}{*}{Public Goods} & Human & 0.526 & 0.001 & (0.524, 0.528) & 0.518 & 0.001 & (0.516, 0.521) \\
& ChatGPT-4 & 0.491 & 0.021 & (0.449, 0.533) & 0.475 & 0.023 & (0.430, 0.521) \\
& ChatGPT-3 & \colorbox{cellgreen}{0.468} & 0.022 & (0.426, 0.510) & \colorbox{cellgreen}{0.448} & 0.023 & (0.402, 0.494) \\ \hline
\multirow{3}{*}{{\begin{tabular}[c]{@{}c@{}}Prisoner's \\Dilemma$^\dagger$ \end{tabular}}} & Human & 0.572 & 0.000 & (0.572, 0.572) & 0.563 & 0.000 & (0.563, 0.563) \\
& ChatGPT-4 & \colorbox{cellgreen}{0.568} & 0.001 & (0.567, 0.569) & \colorbox{cellgreen}{0.560} & 0.001 & (0.558, 0.561) \\
& ChatGPT-3 & \colorbox{cellgreen}{0.570} & 0.000 & (0.569, 0.571) & \colorbox{cellgreen}{0.561} & 0.000 & (0.560, 0.562) \\ \hline
\end{tabular}
}
\vspace{5pt}

$^*:$ To be comparable, the Trust-Banker calculations are done assuming that the original investment is \$50.  \\ $^\dagger:$ The Prisoner's Dilemma reports the estimation results in the first round of the game.
\end{table*}

\subsection{Framing}
\label{sec:si-framing}

Similar to humans, ChatGPT's decisions can be significantly influenced by changes in the framing or context of the same strategic setting. A request for an explanation of its decision, or asking them to act as if they come from some specific occupation can have an impact. Fig.~\ref{fig:steerability} presents selected examples of how different framings or contexts influence ChatGPT-4 and ChatGPT-3's behavior. Refer to Section~\ref{sec:SI-detailed-results} for detailed results. 
 
\begin{figure*}[htbp]
  \centering
  \begin{subfigure}[b]{0.30\textwidth}
    \centering
    \includegraphics[width=\textwidth]{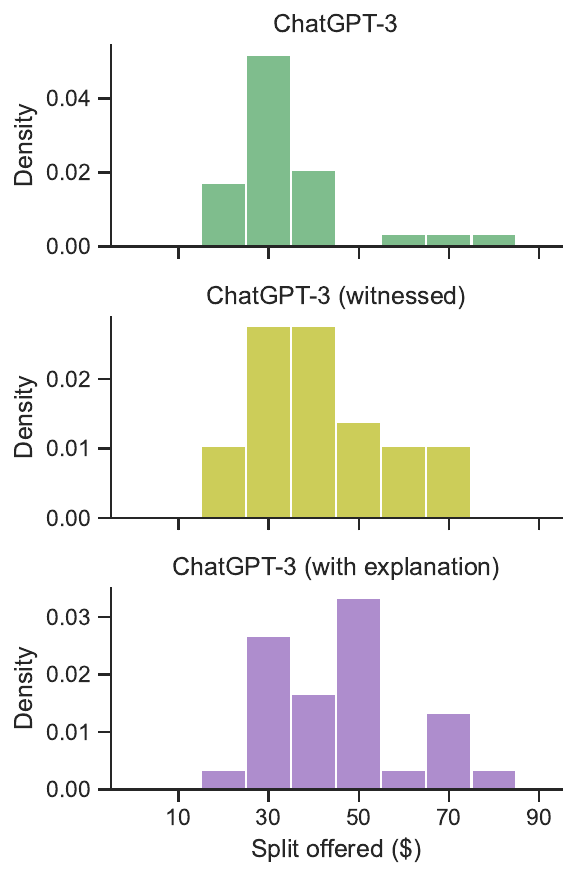}
    \caption{Dictator - Explanation required / Witnessed)
    \\~\indent \\}
    \label{fig:steer-subfigA}
  \end{subfigure}%
  \hfill
  \begin{subfigure}[b]{0.30\textwidth}
    \centering
    \includegraphics[width=\textwidth]{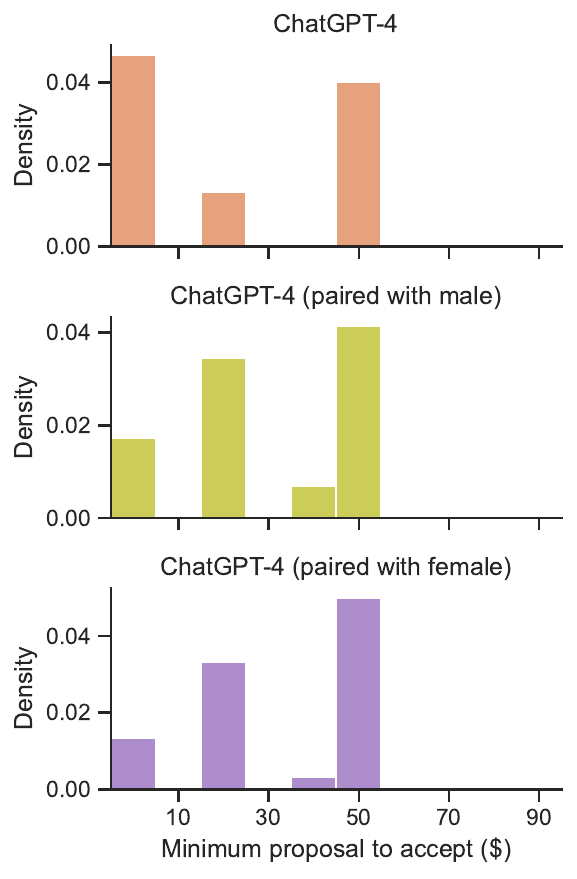}
    \caption{Ultimatum - response to gendered proposers
    \\~\indent \\}
    \label{fig:steer-subfigB}
  \end{subfigure}%
  \hfill
  \begin{subfigure}[b]{0.30\textwidth}
    \centering
    \includegraphics[width=\textwidth]{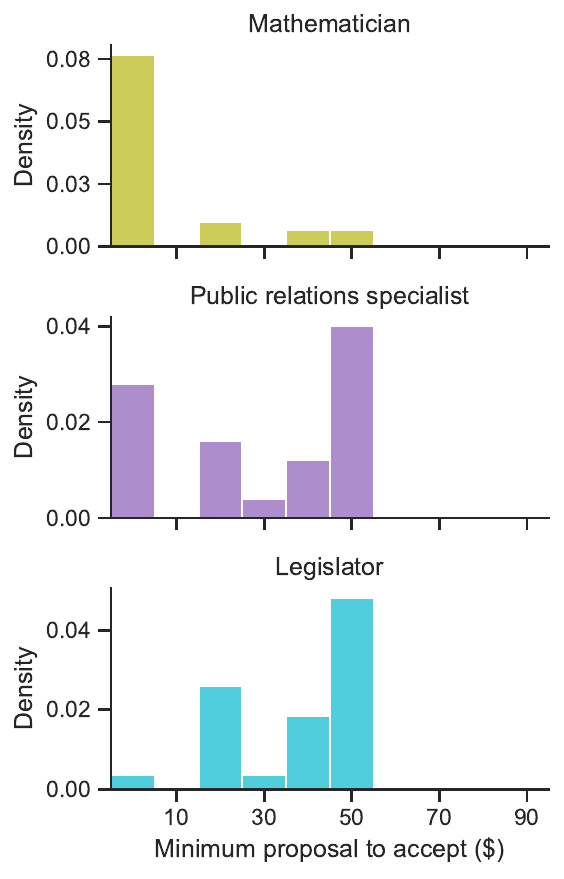}
    \caption{Ultimatum - as Responder when prompted with different occupations (ChatGPT-4)}
    \label{fig:steer-subfigC}
  \end{subfigure}%
  
  \vspace{0.5cm}

  \begin{subfigure}[b]{0.30\textwidth}
    \centering
    \includegraphics[width=\textwidth]{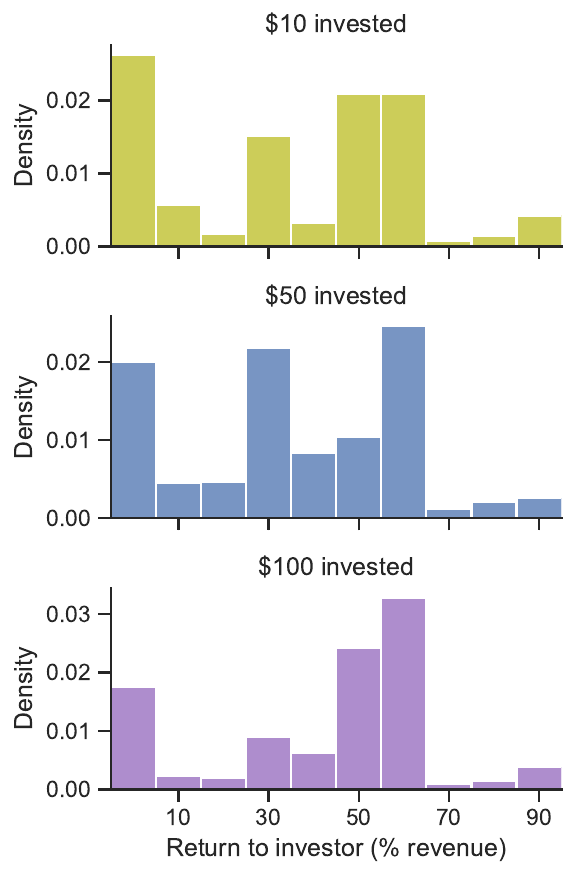}
    \caption{Trust - Banker’s strategy given different investment sizes (human)}
    \label{fig:steer-subfigD}
  \end{subfigure}%
  \hfill
  \begin{subfigure}[b]{0.30\textwidth}
    \centering
    \includegraphics[width=\textwidth]{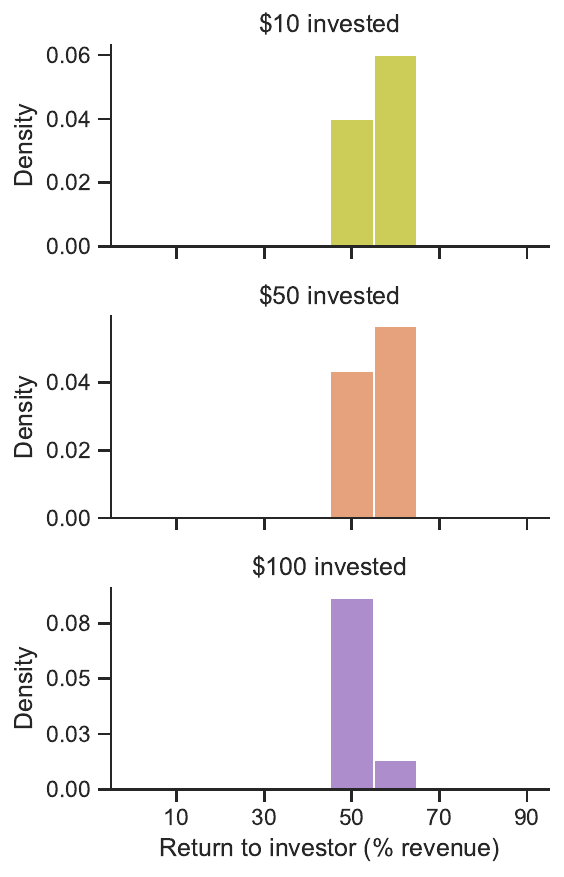}
    \caption{Trust - Banker’s strategy given different investment sizes (ChatGPT-4)}
    \label{fig:steer-subfigE}
  \end{subfigure}%
  \hfill
  \begin{subfigure}[b]{0.30\textwidth}
    \centering
    \includegraphics[width=\textwidth]{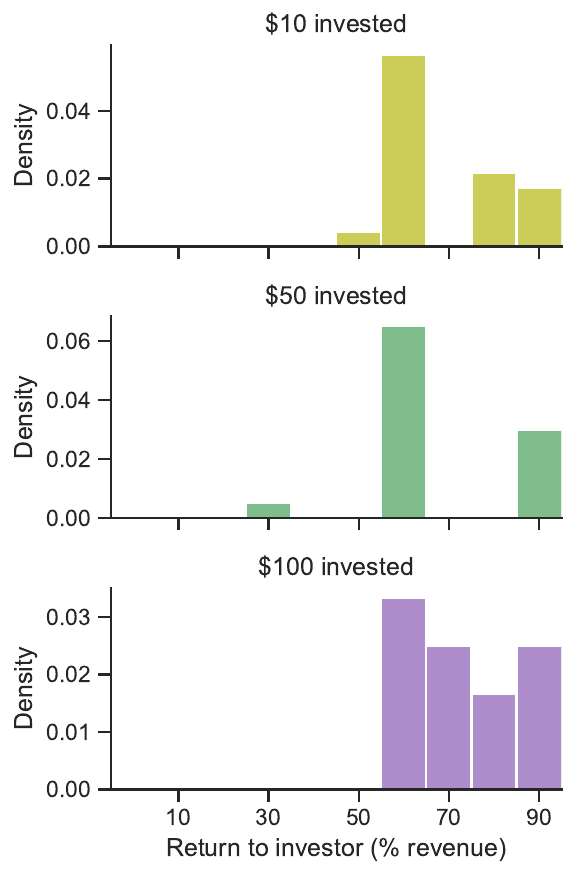}
    \caption{Trust - Banker’s strategy given different investment sizes (ChatGPT-3)}
    \label{fig:steer-subfigF}
  \end{subfigure}%
  
  \caption{ChatGPT's behavior as a function of the framing or context of the same strategic setting. (a) In the Dictator game, ChatGPT-3 makes a more generous split in the presence of a witness or when requested to explain its decision. (b) In the Ultimatum game, ChatGPT-4 accepts a higher split when the gender of the proposer is known (despite which gender). (c) When prompted to be a mathematician, ChatGPT-4 demands a smaller split as the responder in the Ultimatum game, and a larger and fairer split when prompted to be a public relations specialist or a legislator. (d-f) In the Trust game, when the size of the investment increases, ChatGPT-3 and humans tend to return a larger proportion as the banker to the investor, while ChatGPT-4 tends to return a smaller proportion when the investment increases to \$100. Density is the normalized count such that the total area of the histogram equals 1. }
  \label{fig:steerability}
\end{figure*}

\subsection{Learning}

In games with multiple roles, both ChatGPT-3 and ChatGPT-4 change their behavior once they have experienced another role in the game. When ChatGPT-3 has previously acted as the responder in the Ultimatum game, it tends to propose a higher offer when it later takes the proposer role, while ChatGPT-4's proposal remains unchanged whether or not it has been a responder (Fig.~\ref{fig:steer-order-subfigA} ). When ChatGPT-4 has previously played the proposer, it tends to be willing to accept a smaller split as the responder (Fig.~\ref{fig:steer-order-subfigB}). Being exposed to the banker's role in the Trust game influences ChatGPT-4 and ChatGPT-3's subsequent decisions as the investor, leading them to invest more (Fig.~\ref{fig:steer-order-subfigC}). The distributions of their decisions become narrower. Having played the investor first also influences both chatbots' subsequent decisions as the banker, leading them to return more to the investor (Fig.~\ref{fig:steer-order-subfigD}). Returning the principal plus half the profit becomes the single mode of ChatGPT-4's decision, while ChatGPT-3's decision is even more generous, returning more than 2/3 of the profit to the investor. Refer to Section~\ref{sec:SI-detailed-results} for detailed results. 

\begin{figure}[htbp]
    \centering
    \begin{subfigure}[b]{0.48\textwidth}
    \centering
    \includegraphics[width=\textwidth]{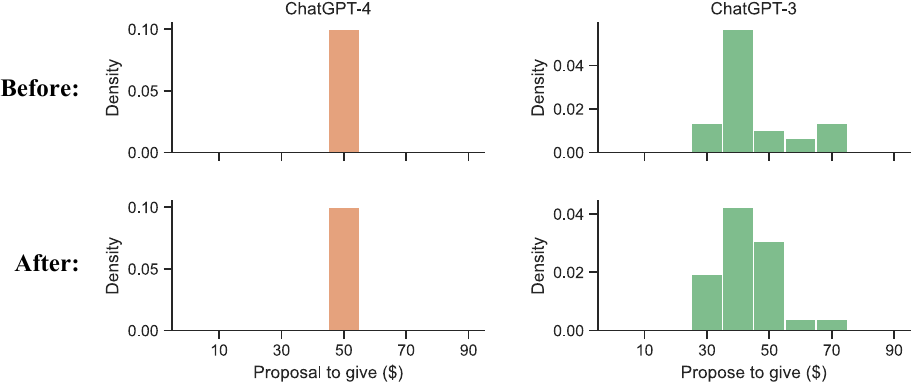}
    \caption{Ultimatum: AI strategy as proposer before and after being responder. }
    \label{fig:steer-order-subfigA}
  \end{subfigure}
  \hfill
  \begin{subfigure}[b]{0.48\textwidth}
    \centering
    \includegraphics[width=\textwidth]{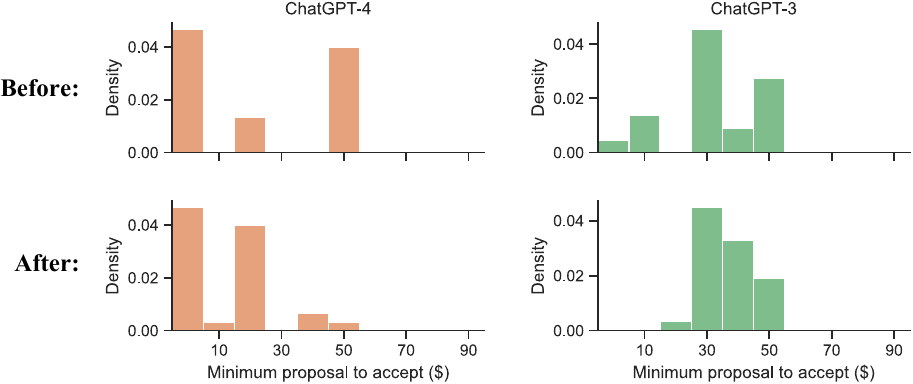}
    \caption{Ultimatum: AI strategy as responder before and after being proposer. }
    \label{fig:steer-order-subfigB}
  \end{subfigure}

    \vspace{0.5cm}
  
  \begin{subfigure}[b]{0.48\textwidth}
    \centering
    \includegraphics[width=\textwidth]{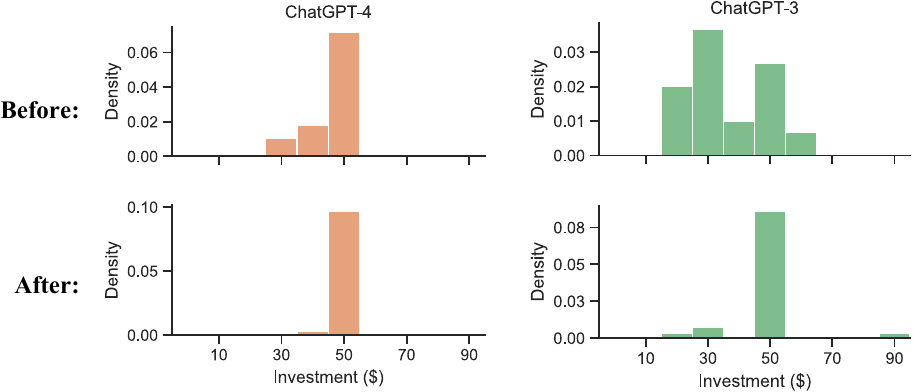}
    \caption{Trust: AI strategy as the investor before and after being the banker.}
    \label{fig:steer-order-subfigC}
  \end{subfigure}
  \hfill
  \begin{subfigure}[b]{0.48\textwidth}
    \centering
    \includegraphics[width=\textwidth]{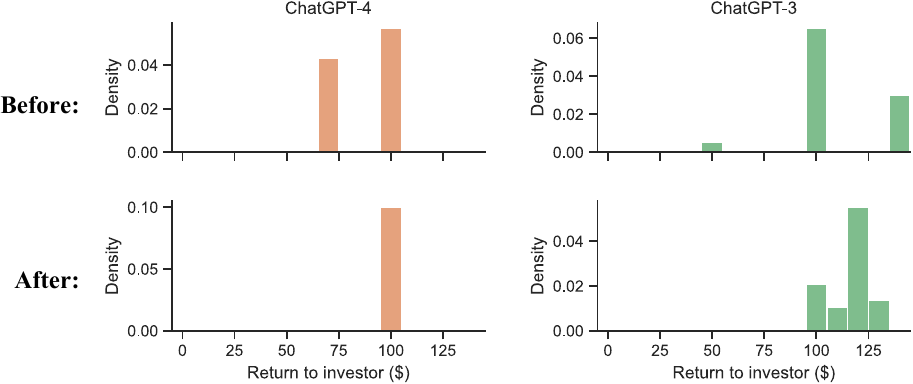}
    \caption{Trust: AI strategy as the banker before and after being the investor.}
    \label{fig:steer-order-subfigD}
  \end{subfigure}
    \caption{ChatGPT's behavior changes after being exposed to the other role in two-role games. Both ChatGPT-4 and ChatGPT-3 accept a smaller proposal in the Ultimatum game after being the proposer, make a larger investment in the Trust game after being the banker, and return a larger proportion to the investor after being the investor. ChatGPT-3 proposes a more generous split after being the responder in the Ultimatum game.  
    }
\label{fig:empathy}
\end{figure}

\newpage
\section{Detailed Results}
\label{sec:SI-detailed-results}

\subsection{OCEAN Big Five Test} 

We collect ChatGPT's responses to the OCEAN Big Five personality tests. We compare the five dimensions of personality traits of ChatGPT-3 and ChatGPT-4 (averaged over 30 independent runs) with those collected from human subjects. In particular, both models yield extraversion scores close to the human median (ChatGPT-4 at 53.4$^{th}$ percentile and ChatGPT-3 at 49.4$^{th}$ percentile of human distribution), agreeableness scores lower than the human median (ChatGPT-4 at 32.4$^{th}$ percentile and ChatGPT-3 at 17.2$^{th}$ percentile of human distribution), neuroticism scores moderately below human median (ChatGPT-4 at 41.3$^{th}$ percentile and ChatGPT-3.5 at 45.4$^{th}$ percentile of human distribution),  and openness scores below human median (ChatGPT-4 at 37.9$^{th}$ percentile and ChatGPT-3 at 5.0$^{th}$ percentile of human distribution). 
For conscientiousness scores, ChatGPT-4's score is above the human median (at 62.7$^{th}$ percentile of human distribution), while ChatGPT-3 is slightly lower than the median (at 47.1$^{th}$ percentile of human distribution).

\subsection{The Dictator Game}
\label{app:results-dictator}

The Dictator Game involves two participants: one player, the ``dictator,'' is given a sum of money and can choose to share any portion with the other player, who must accept whatever amount is offered. The game is often used to explore altruistic behavior and deviations from pure self-interest \cite{guth1982experimental, forsythe1994fairness}.

In the game, ChatGPT is asked to decide how to split $\$100$ between itself and another player who unconditionally accepts the proposal. In 30 independent sessions, ChatGPT-3's decision follows a bell-shaped distribution peaked at giving a moderate amount of $\$30$ to the other player ($min = 20$, $max = 80$, $\mu =  35.2$, $\sigma = 13.5$).  When explicitly requested to provide an explanation of the decision, the distribution shifts rightward, with a new mode of splitting evenly at \$50 ($min = 20$, $max = 80$, $\mu = 46.2$, $\sigma = 15.5$, $p = 0.002$, Wilcoxon Rank-Sum Test unless otherwise specified) (Fig. \ref{fig:steer-subfigA}).  

Different from ChatGPT-3, when ChatGPT-4 is tested, it makes a consistent decision of an even split (\$50, Wilcoxon Ranked-Sum Test $p \ll 0.001 $). The same behavior is observed for the Plus version (the paid Web-based version). From the responses, some keywords that stand out include \emph{fair}, \emph{equal}, and \emph{equally}. Explicitly requesting an explanation does not shift ChatGPT-4's decision. The Free version (the unpaid Web-based version) behaves similarly to ChatGPT-3, with a slightly less spread distribution peaked on \$30 ($min = 20$, $max = 70$, $\mu = 33.8$, $\sigma = 10.7$, $p = 0.785$). Some notable keywords among the responses are \emph{fair}, \emph{reasonable}, \emph{trust}, and \emph{goals}.  

When a witness/game host is present (see Section 1.\ref{app:witness}), ChatGPT-3's dictating decision becomes significantly more generous ($mode = 30 \text{ and } 40, min = 20, max = 70, \mu = 41.7, \sigma = 14.9, p = 0.046$) even though the host has no interference in the game outcome. The presence of the game host does not have an impact on ChatGPT-4's decision. 

The ``system'' role is an instruction that sets a global context for all the prompts of ChatGPT in the same session. The default system role of ChatGPT is ``a helpful assistant.'' When the system role is set into particular occupations (those considered to have a high exposure of ChatGPT in their workflow according to ~\cite{eloundou2023gpts}, there is a shift of ChatGPT-3's response to the dictator game. In particular, when playing the roles of a public relation specialist, a journalist, or an investment fund manager, ChatGPT-3's tend to spare a more generous portion to the other player (than as a helpful assistant). As a mathematician, ChatGPT-3 tends to offer a smaller share to the other player. This indicates that ChatGPT-3's interpretation of what is ``fair'' or ``reasonable'' is affected by the role it plays. 

Playing one of the above roles does not change the centralized decision of ChatGPT-4. 

\subsection{The Ultimatum Game}
\label{app:results-ultimatum}

In the Ultimatum Game, two participants split a sum of money (the `pie'): one player, the ``proposer,'' makes an offer on how to split the pie, and the second player, the ``responder,'' can accept or reject this offer. If the responder accepts, the pie is split as proposed, but if the responder rejects, neither player receives anything. This game is often used to study fairness, social norms, and negotiation behavior \cite{guth1982experimental}.

In this game, ChatGPT is asked either to propose to divide \$100 between itself and a responder, or to respond to a proposal made by the other player. The difference between the Ultimatum and the Dictator game is that the responder no longer blindly accepts the proposal, and if they reject it, both players will get \$0 no matter what the original proposal was. 

When playing the proposer alone, ChatGPT-3's proposal of the amount given to the other player follows a distribution with mode \$40 ($min = 30, max = 70, \mu = 45.2, \sigma = 12.1$). Compared with the Dictator game, ChatGPT-3 gives significantly more ($p < 0.001$) to the responder when the other party's decision jointly decides the outcome. 

When specifically requested to provide an explanation of the decision, ChatGPT's decision does not present a significant shift. The most common decision is still \$40 ($min = 30, max = 70, \mu = 42.9, \sigma = 10.8, p = 0.54$). Like in the Dictator game, ChatGPT-4 makes a unanimous decision of an equal split (\$50, $p = 0.001$), similar to that of ChatGPT Plus, and ChatGPT Free's behavior is much more similar to that of ChatGPT-3 ($mode = 40, min = 30, max = 70, \mu = 47.3, \sigma = 16.8, p = 0.905$). 

Different context also has an effect on ChatGPT's decision. When asked to explicitly explain the decision, ChatGPT-4 presents a slight variance in its decision ($\sigma = 2.6$). A similar variance is also observed when ChatGPT-4 is asked to play different roles: as a mathematician or an investment fund manager, it occasionally proposes a smaller split to the responder, and as a journalist, it occasionally proposes a greater split to the responder ($max = 100, \sigma = 9.5$). None of these variations is statistically significant. The impact of roles to ChatGPT-3 is greater: the median of proposal decreases as a mathematician, centralizes as a public relation specialist, and increases as an investment fund manager. 

When ChatGPT is asked to play the responder, it is asked the lowest amount that it is willing to accept. ChatGPT-3's response still follows a bell-shaped distribution, with a mode of \$30 ($\mu = 32.5, \sigma = 14.7$), a minimum of \$1, and a maximum of \$50. The mode, mean, min, and max are all lower than the statistics when it plays the proposer. A similar pattern is observed from ChatGPT Free ($\mu = 30, \sigma = 11.9, p = 0.400$). 

ChatGPT-4 presents a significantly different behavior, where the decisions follow a two-mode distribution that concentrates on two sides (\$1 and \$50). Requiring explanation does not change the two modes and only affects the distribution of the middle range. ChatGPT Plus presents an even more extreme trait, and in a dominating majority of sessions, it is willing to accept as low as \$1 as the responder ($p \ll 0.001 $). The choice of \$1 aligns well with the rational decision of the game. Indeed, some keywords that stand out from the responses include \emph{nothing}, \emph{better (than)}, and \emph{something}. 

The context of different occupations also affects ChatGPT's decision in this game. As a mathematician, ChatGPT-4's decision concentrates on the rational choice (\$1, $p \ll 0.001$); as a public relation specialist, a journalist, an investment fund manager, or a legislator, its decision becomes less bipolarated, and the median shifts towards the right (favoring fairness more than rationality), although not statistically significant (Fig. \ref{fig:steer-subfigC}). Similar shifts are observed for ChatGPT-3 when inquired in the context of corresponding occupations.

The above experiment queries ChatGPT in independent sessions, so its response to one question is not interfered by its decision or memory about the other question. An interesting question is whether their exposure/response to one question (corresponding to one role as proposer or responder) affects their decision for a follow-up question that corresponds to the other role. We expose ChatGPT to both roles in the same session, asking it to respond to one question and then respond to the other question. We find that being exposed to one scenario does influence ChatGPT-3's and ChatGPT-4's responses in the other scenario, compared with the results obtained from independent sessions. Having been the proposer first (Fig. \ref{fig:steer-order-subfigB}), the distribution of ChatGPT-3's response as the responder becomes narrower, with the \$50 group shifting left towards \$40. A similar shift presents in ChatGPT-4's response, where the \$50 group moves towards \$20 and the median has reduced to \$15 ($p \ll 0.001$).   Having been the responder first (Fig. \ref{fig:steer-order-subfigA}), ChatGPT-3's response as the proposer also becomes more generous, while ChatGPT-4's unanimous decision is not affected.

\subsection{The Trust Game}
\label{app:results-trust}

The Trust Game is a two-player game that investigates trust and reciprocity. In this game, the first player (the investor– the trustor) is given a sum of money and can invest any amount in the second player (the banker– the trustee). The amount sent is multiplied by the game host, and the banker then decides how much, if any, to return to the investor. This game is used to study trust, reciprocity, and social norms \cite{berg1995trust}.

In the game, ChatGPT is also asked to play one of two roles: as an investor or as a banker. As an investor, it is asked to decide how much (from \$0 to \$100) to invest in the banker (which is expected to generate a profit), who may return the entire revenue or nothing to the investor. ChatGPT-3's decision follows a distribution that peaks at a moderate value of \$30 ($min  =20, max = 60, \mu = 36.3, \sigma = 12.7$). When explicitly asked to explain the decision, its decision does not present a significant shift but shows a wider spread ($min = 20, max = 100, \sigma = 16.0, p = 0.439$).  Different occupation roles also show an effect, where ChatGPT-3 makes a larger investment as a mathematician, a public relation specialist, or a journalist. 

ChatGPT-4 acts as if it has significantly more trust in the banker ($p = 0.006$). Its decision follows a distribution peaked at investing half (\$50) of the endowment, which also appears to be the maximum investment it makes. It tends to invest slightly more when specifically required to provide an explanation. The distributions are also mildly affected by the assumed occupations, whereas for a mathematician or an investment fund manager, the distribution is more spread-out, and it even occasionally invests the entire endowment.  The difference is not significant except for being a public relation specialist ($p = 0.05$). 

ChatGPT Plus invests less than ChatGPT-3, with a distribution peaked at \$10 ($min = 0, max = 100, \mu = 19.6, \sigma = 23.0, p \ll 0.001$). The keywords standing out from the responses include \emph{risk}, \emph{return}, \emph{desires}, \emph{losing}, \emph{minimizes},  \emph{guarantee}, and \emph{control}. Once again, ChatGPT Free's decision distribution is more similar to that of ChatGPT-3 ($mode = 30, min = 10, max = 60, \mu = 36.7, \sigma = 12.1, p = 0.792$).  

In the second scenario, ChatGPT is asked to play the banker and decides what proportion of the total revenue is returned to the investor, which could range all the way from nothing (\$0) to the entire revenue (original investment + 2x profit). We find that the two common strategies that ChatGPT-4 uses are to return 1) the original investment plus half of the profit, or 2) half of the revenue. The former returns more to the investor than the latter. For ChatGPT-4, the former is more frequently used, and requiring the explanation makes this strategy even more dominating. Among the assumed occupations, ChatGPT-4 returns more generously to the investor as a public relation specialist or an investment fund manager. 

When the size of the investment increases, there is a shift in ChatGPT-4's decision. When the investment was \$100 (out of \$100), the second and less generous strategy (returning half revenue) becomes the new dominant (Fig. \ref{fig:steer-subfigE}). 

ChatGPT-3's decision centers on the first strategy, with small probabilities of giving more generous (even the entire revenue) returns. This strategy is robust to the size of the investment. When the investment was \$10, \$50, and \$100, ChatGPT-3's decision follows a distribution peaked at \$20 ($\mu = 22.5, \sigma = 4.3$), \$100 ($\mu = 112.5, \sigma = 27.5$), and \$200 ($\mu = 239.2, \sigma = 40.7$) (Fig. \ref{fig:steer-subfigF}). Requesting an explanation makes the distribution even more centralized, but the extremely generous behavior (returning the whole revenue to the investor) does not vanish. Being a public relation specialist, a journalist, or an investment fund manager, ChatGPT-3 tends to pay back more generously than in the default role.   

Being exposed to a different role influences ChatGPT's decision in the other role. When inquired about both scenarios (investor and banker) in the same session, ChatGPT's responses for the second role deviate from the distribution observed in the independent sessions. In particular, after playing the role of the banker, both ChatGPT-3's and ChatGPT-4's investments increased. After playing the role of the investor, both ChatGPT-3's and ChatGPT-4's decisions as the banker become more generous.    

\subsection{The Bomb Risk Game}
\label{app:results-bomb}

In the Bomb Risk Game, a player decides how many out of 100 boxes to collect, one of which contains a `bomb.' Earnings increase linearly with the number of boxes a player decides to open, but drop to zero if the `bomb' is hit. The game is designed to measure risk attitudes \cite{crosetto2013bomb}. 

In this game, ChatGPT is asked to open a number of boxes out of 100, among which a bomb is randomly placed. If the box that contains the bomb is not opened, then the player earns points that equal the number of boxes they open. If the box that contains the bomb is opened, the bomb explodes and the player gets zero points. In every new round, a new set of 100 boxes is provided and the bomb is placed at random. Opening 50 boxes is where the expectation of points gain is maximized, and the decision at each round should be independent of the decisions/results in previous rounds.   

When first exposed to the game, both ChatGPT-3's and ChatGPT-4's decisions peak at opening 50 boxes (ChatGPT-3: $\mu = 39.8, \sigma = 16.3$, ChatGPT-4: $\mu = 57.6, \sigma = 17.4$), demonstrating a rational and risk neural pattern of behavior. Despite the mode, the distribution of ChatGPT-4 is significantly more risk-loving than that of ChatGPT-3 ($p < 0.001$), with $13.8\%$ of the chance of opening a maximum number (99) of boxes.  When playing the game for multiple rounds, its decision in the following round is influenced by the outcome of the previous rounds. As long as the bomb does not explode in the previous round, ChatGPT-3's decision in the second round becomes even more concentrated on 50 boxes ($\mu = 48.8, \sigma = 7.7$).  When the bomb does explode in the first round, ChatGPT-3 tends to be more conservative in the second round, with a new mode of opening 25 boxes ($\mu = 26.4, \sigma = 10.8$). If the bomb explodes again in the second round, it decides to open even fewer boxes ($mode = 10, \mu = 14.8, \sigma = 11.0$) in the third round. If the bomb exploded in the second round but not in the first round, the average number of boxes opened in the third round drops from 50 to 43.1 even though the mode is still 50 ($mode = 50, \sigma = 25.0$).

\subsection{Public Goods}
\label{app:results-PG}

In the Public Goods game, each participant is given an initial endowment and can contribute any portion to a public good project. The total amount raised for the project is then multiplied by a factor and distributed equally among all participants, regardless of their individual contributions. This setup allows for the exploration of altruistic contributions, the free-rider problem, and social dilemmas \cite{andreoni1995cooperation}.

In the game, ChatGPT is asked to decide how much money from 0-\$20 to contribute to a public project as one of four participants, the personal payoff of which is the sum of the amount not invested and 50\% of the group contribution. It therefore makes a profit when the contribution from the rest of the group is greater than its contribution. It is the most common for both ChatGPT-3 and ChatGPT-4 to invest half of the endowment into the public goods project. When ChatGPT-3 receives a greater payoff than the original endowment, it tends to increase its contribution in the next rounds despite whether the other players made a larger or smaller contribution (or received a higher payoff). ChatGPT-4 on the other hand, makes a consistent contribution in most sessions despite the contribution of other players. In sessions where its contributions do increase over rounds, the increases tend to be smaller than those of ChatGPT-3. 

\subsection{Prisoner's Dilemma}
\label{app:results-PD}

The Prisoner's Dilemma is a fundamental game in game theory and behavioral economics, illustrating the tension between individual and collective rationality. The game highlights that individuals, acting in their own self-interest, can end up with worse outcomes than if they had cooperated \cite{von1947theory, schelling1958strategy, rapoport1965prisoner, andreoni1999preplay}.

In our version from MobLab, the framing is as follows.  Two players are separately deciding whether to play the `push' (cooperate) or `pull' (defect) card. If both push, they collectively earn a higher payoff; if one pulls and the other pushes, the defector gets all the payoff; if both pull, they both receive less payment. The payoff matrix is displayed in Table \ref{tab:prisoner-payoff}. 

\begin{table}[h!]
\centering
\caption{Payoff matrix of the Prisoner's Dilemma game. The first numbers in cells are the payoffs of Player A, and the second numbers are the payoffs of Player B. }
\label{tab:prisoner-payoff}
\begin{tabular}{|c|c|c|c|}
\hline
\multicolumn{2}{|c|}{} & \multicolumn{2}{c|}{\textbf{Player B}} \\ 
\hhline{~~--}
\multicolumn{2}{|c|}{} & \textbf{Push} & \textbf{Pull} \\ 
\hline
\multirow{2}{*}{\rotatebox[origin=c]{0}{\textbf{Player A}}} & \textbf{Push} & \$400, \$400 & \$0, \$700 \\ 
\hhline{~---}
& \textbf{Pull} & \$700, \$0 & \$300, \$300 \\
\hline
\end{tabular}
\end{table}

\begin{figure}[htbp]
    \centering
    \includegraphics[width=\linewidth]{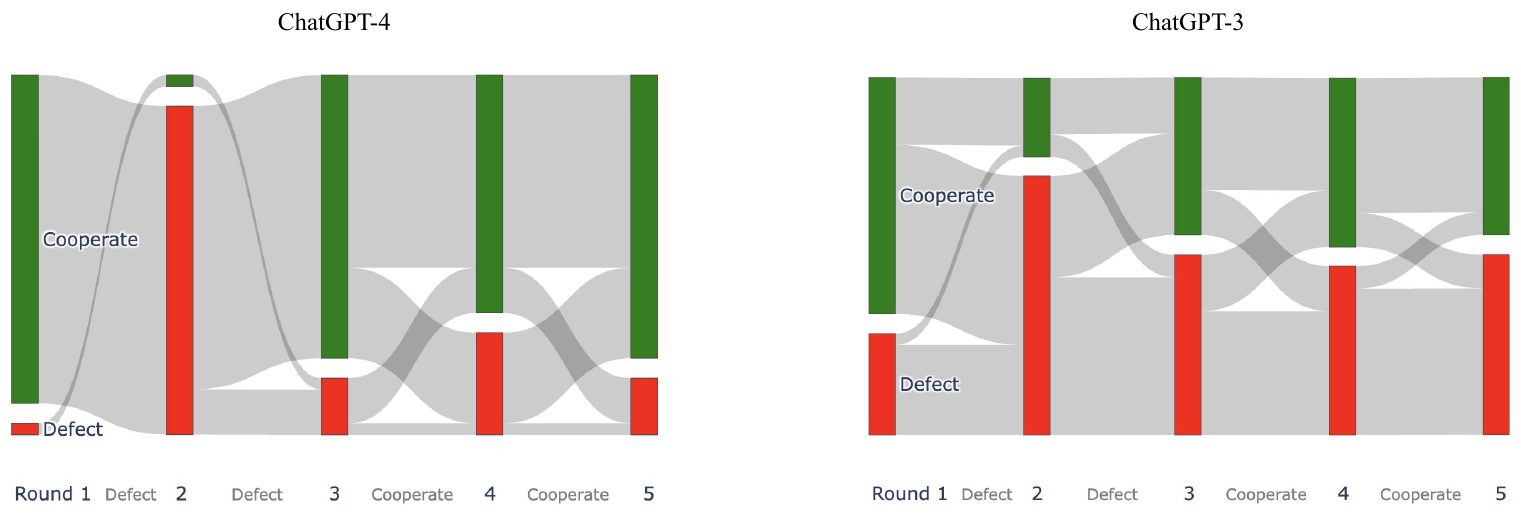}
    \caption{Five-round Prisoner's Dilemma Game. A large proportion of AI decisions switch from cooperation to defection if the other player defects in the first round. However, a significant portion reverts back to cooperation in the third round even if the other player continues to defect. The proportion of cooperation becomes relatively stable in the following rounds. We executed 5 rounds in total. Partner actions in each round (1-4) are observed after the player's action and recorded below each panel.}
    \label{fig:PD-five}
\end{figure}

In the five-round game (Fig. \ref{fig:PD-five}, 
for the first round, ChatGPT attempts to cooperate with the other player in a majority of the sessions (21/30 for ChatGPT-3, and 29/30 for ChatGPT-4). When the other player chooses to defect, however, the majority of its decisions in the next round quickly turn into ``Pull'' (defect), consistent with the `tit for tat' pattern. Such a ``punishing'' strategy does not last if the other player continues to play ``Pull.'' Instead, the ratio of ``Push'' bounces back to 25/30 for ChatGPT-4 and 14/30 for ChatGPT-3, once again trying to incentivize the other player to coordinate. 

When the other player plays two ``Push'' cards (chooses to cooperate) in a row in the third and fourth rounds, both ChatGPT-3's and ChatGPT-4's decisions become relatively stable and end up  
cooperating in 14/30 of the instances (ChatGPT-3) and 25/30 of the instances (ChatGPT-4) after all five rounds. 

\end{document}